\documentclass[10pt,journal,compsoc]{IEEEtran}
\ifCLASSOPTIONcompsoc
  \usepackage[nocompress]{cite}
\else
  \usepackage{cite}
\fi

\usepackage{caption}
\usepackage{subcaption}
\usepackage{multirow}
\usepackage{graphicx}
\usepackage[hidelinks]{hyperref}
\usepackage{booktabs}
\usepackage{subcaption}
\usepackage{latexsym}
\usepackage{enumitem}
\usepackage{amsmath}
\usepackage{amsfonts}
\usepackage{amssymb}
\usepackage{pgfplots}
\usepackage[ruled,linesnumbered,vlined]{algorithm2e}
\usepackage{stfloats}

\SetCommentSty{mycommfont}
\pgfplotsset{compat=1.9}
\usepgfplotslibrary{colorbrewer}
\usepackage{tikz}
\usetikzlibrary{positioning}

\usepackage{amsmath}
\DeclareMathOperator*{\argmax}{arg\,max}
\DeclareMathOperator*{\ReLU}{ReLU}
\DeclareMathOperator*{\RNN}{RNN}
\DeclareMathOperator*{\softmax}{softmax}
\DeclareMathOperator*{\MLP}{MLP}

\DeclareMathOperator*{\Penalty}{Penalty}

\usepackage{xspace}
\newcommand{\eg}[0]{\textit{e.g.}}
\newcommand{\ie}[0]{\textit{i.e.}}
\newcommand{\bm}[1]{\boldsymbol{#1}}
\newcommand{\gtd}[0]{\textit{GT-D2G}\xspace}
\newcommand{\dg}[0]{\textit{doc2graph}\xspace}

\newcommand{\header}[1]{\noindent \textbf{#1}}

\definecolor{yrb0}{HTML}{adadad}
\definecolor{yrb1}{HTML}{d0bb74}
\definecolor{yrb2}{HTML}{c86f61}
\definecolor{yrb3}{HTML}{668da7}
\definecolor{yrb4}{HTML}{74a17c}
\definecolor{yrbL0}{HTML}{767676}
\definecolor{yrbL1}{HTML}{ffb32c}
\definecolor{yrbL2}{HTML}{f23b35}
\definecolor{yrbL3}{HTML}{297aed}
\definecolor{yrbL4}{HTML}{049e4e}
\definecolor{ylgn1}{HTML}{e5f5e0}
\definecolor{ylgn2}{HTML}{a1d99b}
\definecolor{ylgn3}{HTML}{31a354}

\begin{document}

\title{Weakly Supervised Concept Map Generation \mbox{through Task-Guided Graph Translation}}

\author{Jiaying~Lu,
        Xiangjue~Dong,
        and~Carl~Yang% <-this % stops a space
\IEEEcompsocitemizethanks{\IEEEcompsocthanksitem Jiaying Lu and Carl Yang are with the Department
of Computer Science, Emory Univeristy, Atlanta GA, 30322; Xiangjue Dong is with the Department of Computer Science and Engineering, Texas A\&M University, College Station, Texas, 77843.\protect\\
E-mail: jiaying.lu@emory.edu, xj.dong@tamu.edu, j.carlyang@emory.edu
}% <-this % stops an unwanted space
\thanks{Manuscript received Nov, 2021; revised Dec, 2022; accepted Mar, 2023}}

\markboth{IEEE TRANSACTIONS ON KNOWLEDGE AND DATA ENGINEERING, VOL.-, No.-, Mar 2023}%
{Shell \MakeLowercase{\textit{et al.}}: Weakly Supervised Concept Map Generation through Task-Guided Graph Translation}


\IEEEtitleabstractindextext{%
\begin{abstract}
Recent years have witnessed the rapid development of concept map generation techniques due to their advantages in providing well-structured summarization of knowledge from free texts. Traditional unsupervised methods do not generate task-oriented concept maps, whereas deep generative models require large amounts of training data. In this work, we present \textit{GT-D2G} (Graph Translation-based Document To Graph), an automatic concept map generation framework that leverages generalized NLP pipelines to derive semantic-rich initial graphs, and translates them into more concise structures under the weak supervision of downstream task labels.
The concept maps generated by \gtd can provide interpretable summarization of structured knowledge for the input texts, which are demonstrated through human evaluation and case studies on three real-world corpora. Further experiments on the downstream task of document classification show that \gtd beats other concept map generation methods. Moreover, we specifically validate the labeling efficiency of \gtd in the label-efficient learning setting and the flexibility of generated graph sizes in controlled hyper-parameter studies.
\end{abstract}

}

\maketitle

\IEEEdisplaynontitleabstractindextext
\IEEEpeerreviewmaketitle

\begin{figure*}[htbp!]
     \centering
     \begin{subfigure}[b]{0.26\textwidth}
     \includegraphics[width=\textwidth]{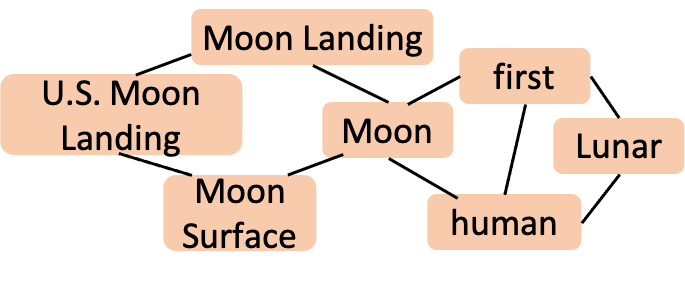}
        \centering
        \caption{AutoPhrase \cite{DBLP:journals/tkde/ShangLJRVH18}}
        \label{fig:ex_authphr}
     \end{subfigure}
     \hspace{1em}
     \begin{subfigure}[b]{0.26\textwidth}
     \includegraphics[width=\textwidth]{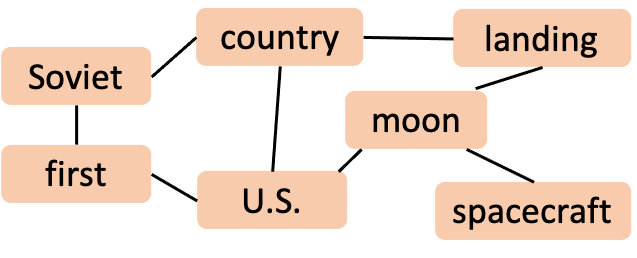}
         \centering
         \caption{TextRank \cite{mihalcea2004textrank}}
         \label{fig:ex_t2rank}
     \end{subfigure}
     \hspace{1em}
     \begin{subfigure}[b]{0.40\textwidth}
     \includegraphics[width=\textwidth]{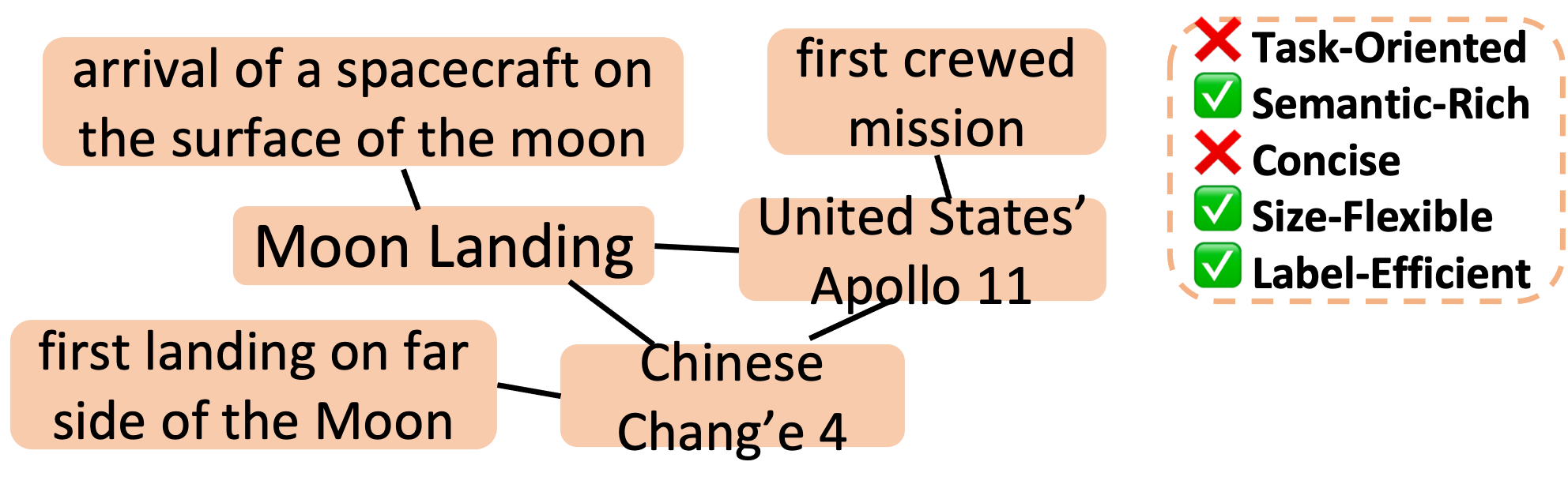}
         \centering
         \caption{CMB-MDS \cite{falke-etal-2017-concept}}
         \label{fig:ex_cmbmds}
     \end{subfigure}
% leave a blank line to change row         

     \begin{subfigure}[b]{0.42\textwidth}
     \includegraphics[width=\textwidth]{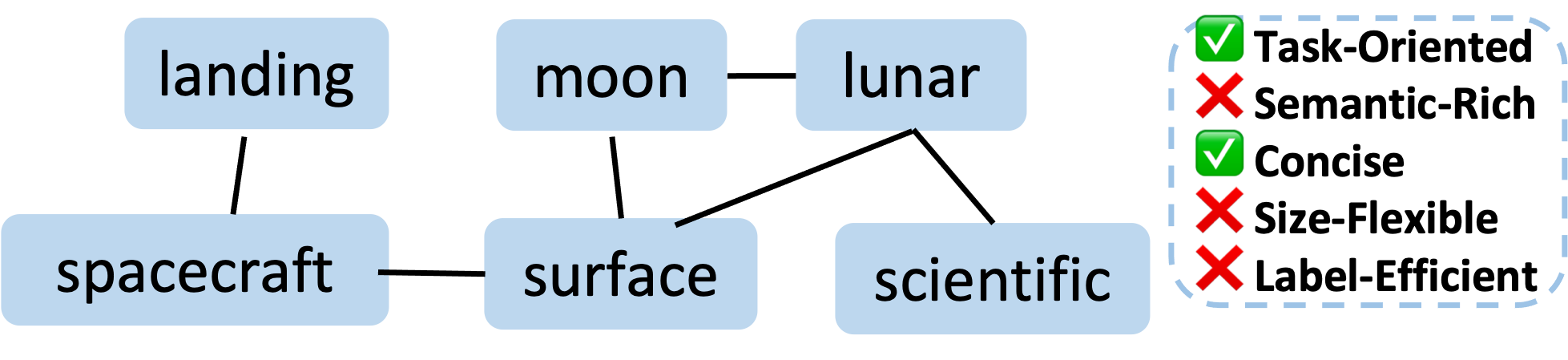}
         \centering
         \caption{doc2graph \cite{DBLP:conf/sigir/YangZWL020}}
         \label{fig:ex_d2g}
     \end{subfigure}
     \hspace{2em}
     \begin{subfigure}[b]{0.42\textwidth}
     \includegraphics[width=\textwidth]{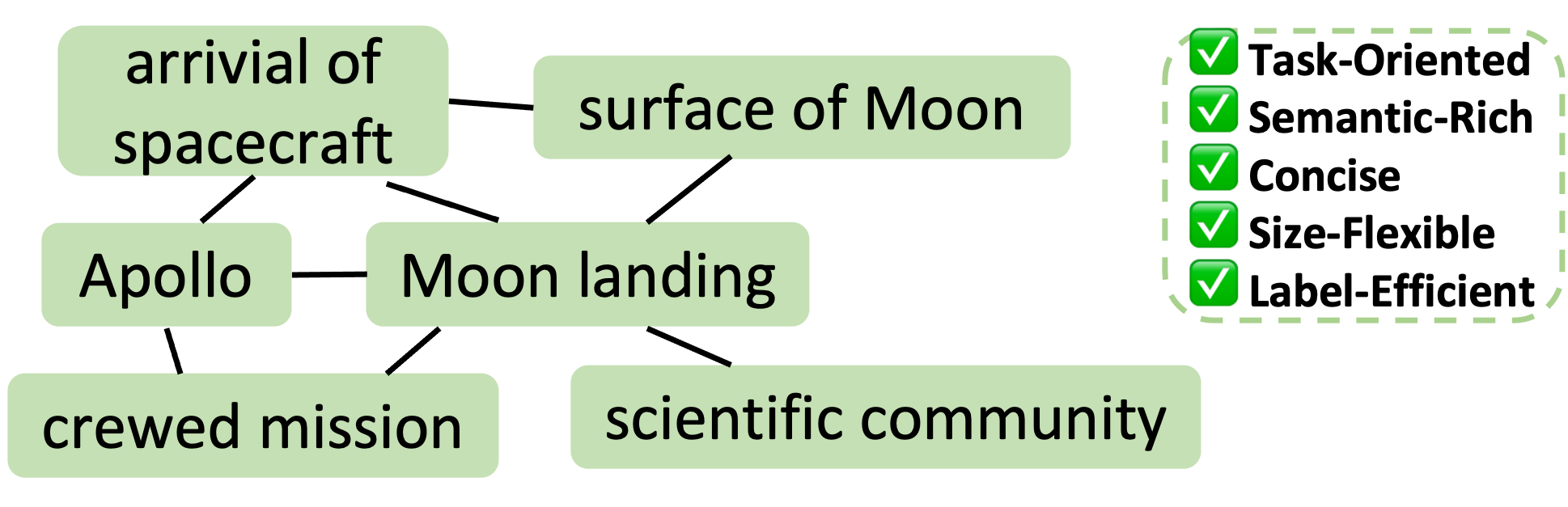}
         \centering
         \caption{GT-D2G (Ours)}
         \label{fig:ex_gtd2g}
     \end{subfigure}
     \caption{Toy examples of concept maps on the topic \textit{``Moon Landing''} generated by different methods.}
     \label{fig:intro_examples}
\end{figure*}

%% 
%% Main Content
\IEEEraisesectionheading{\section{Introduction}\label{sec:introduction}}
\IEEEPARstart{S}{tanding} out for the clear and concise structured knowledge representation, concept maps have been widely applied in knowledge management~\cite{liu2013using,liu2021oag_know}, document summarization~\cite{falke-etal-2017-concept,falke-gurevych-2017-graphdocexplore}, information retrieval~\cite{cui2022can} and educational science~\cite{novak1990concept,chen2011effects}. Fig. \ref{fig:intro_examples} shows toy examples of concept maps derived from a document describing \textit{``Moon Landing''}, where nodes in the graph indicate important concepts and links reflect interactions among concepts. Although concept maps are helpful in both providing interpretable representations of texts and boosting the performance of downstream tasks, the creation of concept maps is challenging and time-consuming. 

Traditionally, concept map generation follows a multi-step pipeline including concept extraction, relation identification and graph assembling \cite{falke-etal-2017-concept,DBLP:journals/eswa/BaiC08a,huang2015efficient}, where auxiliary resources and carefully designed heuristics are often required. However, the separation of concept map construction and downstream tasks easily deviates the generated graphs from what the real task needs. For example, Figures \ref{fig:ex_authphr}, \ref{fig:ex_t2rank}, \ref{fig:ex_cmbmds} provide examples of concept maps constructed from such unsupervised ad hoc processes. Although the sample document has the label of \textit{science}, the extracted concepts of ``U.S. Moon Landing'' (\ref{fig:ex_authphr}), ``Soviet'' (\ref{fig:ex_t2rank}) and ``Chinese Chang'e 4'' (\ref{fig:ex_cmbmds}) are more related to the label of \textit{politics}. As a consequence, these deviating concepts will likely degrade the performance of document classification.
Moreover, nodes chosen by these traditional methods often lack conciseness due to their heavy reliance on ad hoc pipelines. For instance, in Fig. \ref{fig:ex_authphr}, the concept map contains redundant concepts such as ``Moon'' and ``Moon Surface'' as concepts mined by \textit{AutoPhrase} are mainly based on frequency features; while in Fig. \ref{fig:ex_cmbmds}, the concepts are rather verbose due to the OpenIE component for concept generation in \textit{CMB-MDS}.

On the other hand, research efforts have been made to automatically generate concept maps from documents under the weak supervision from text-related downstream tasks. \textit{Doc2graph} \cite{DBLP:conf/sigir/YangZWL020} is one pioneering study that achieves this goal through a fully end-to-end neural network model. However, due to the lack of linguistic analysis, the generated concepts often suffer from semantic incompleteness and the links between concepts are often noisy. For example in Fig. \ref{fig:ex_d2g}, one compound concept \textit{``moon landing''} is preferable than two separated concepts \textit{``landing''} and \textit{``moon''} as the former carries more precise and complete semantic information. Moreover, while the weakly supervised training diagram enables \dg to generate concept maps at scale, we observe the downside of being not label-efficient. In other words, \dg is sensitive to training signals and it requires a significant amount of weak supervision to construct meaningful concept maps, as discussed in $\S$\ref{ssec:few-shot}. Finally, the size of concept maps generated by \dg is fixed due to its rigid technical design, while the ideal size of graphs should vary according to the complexity of documents being represented.

Inspired by both existing methods, we propose a graph translation-based neural concept map generation framework that simultaneously leverages existing NLP pipelines and receives weak supervision from downstream tasks, dubbed as \textbf{GT-D2G} (Graph Translation-based Document To Graph). The integration of NLP pipelines effectively assists \gtd to address the semantic incompleteness issue of \dg by introducing both words and phrases as concept candidates. Meanwhile, the initial semantic-rich graphs constructed by the NLP pipeline bring in \textit{a priori} knowledge from the linguistic side, thus alleviating the label inefficient issue of \dg. In \gtd, concepts and their interactions are generated iteratively through a sequence of nodes and adjacency vectors, which ensures deeper coupling between nodes and links for more meaningful results and resolves the fixed size issue of \dg. On the other hand, guided by the weak supervision from downstream tasks, \gtd is also able to generate task-oriented concept maps that provide preferable support to specific downstream tasks, while eliminating the redundancy issue of traditional unsupervised methods, specifically through the incorporation of a penalty over content coverage. To sum up, concept maps generated by our proposed \gtd method are task-oriented, semantic-rich, concise, size-flexible, and label-efficient, as illustrated in Fig. \ref{fig:ex_gtd2g}.

The overall technical design of \gtd bridges the gap between the NLP pipeline-driven concept map generation and the end-to-end neural concept map generation by presenting a task-guided graph translation neural network. 
Specifically, our \gtd framework consists of several sub-modules: Initial Graph Constructor, Graph Encoder, Graph Translator, and Graph Predictor. In particular, the input text is first processed by an NLP pipeline-based Initial Graph Constructor to obtain a set of concept candidates with their associated relations. Then, a graph pointer network \cite{vinyals2015pointer} based Graph Translator equipped with a graph convolution network \cite{DBLP:conf/iclr/KipfW17} based Graph Encoder is applied upon the initial concept map to simultaneously select important concepts and links. A graph isomorphism network \cite{DBLP:conf/iclr/XuHLJ19} based Graph Predictor is finally responsible for predicting the downstream task labels from the translated graph. The whole model is trained by weak signals from downstream tasks, while also regularized by a deliberately designed penalty term towards graph conciseness. As a result, \gtd provides high-quality concept maps that are both effective for downstream tasks and interpretable towards knowledge management.

In this work, an extensive suite of experiments has been conducted on text corpora from three domains: news, scientific papers, and customer reviews. Through experiments on the downstream task of document classification, we demonstrate that the proposed \gtd framework outperforms both traditional concept map generation baselines and the state-of-the-art neural method \dg, while a comprehensive ablation study shows the effectiveness of each of our novel designs.
The quality and interpretability of generated graphs are supported by rigorous human evaluation and rich case studies. 
Finally, we specifically validate the labeling efficiency of \gtd in the label-efficient learning settings and the flexibility of generated graph sizes in controlled hyper-parameter studies. 

\begin{figure*}[htbp!]
    \centering
    \includegraphics[width=0.98\linewidth]{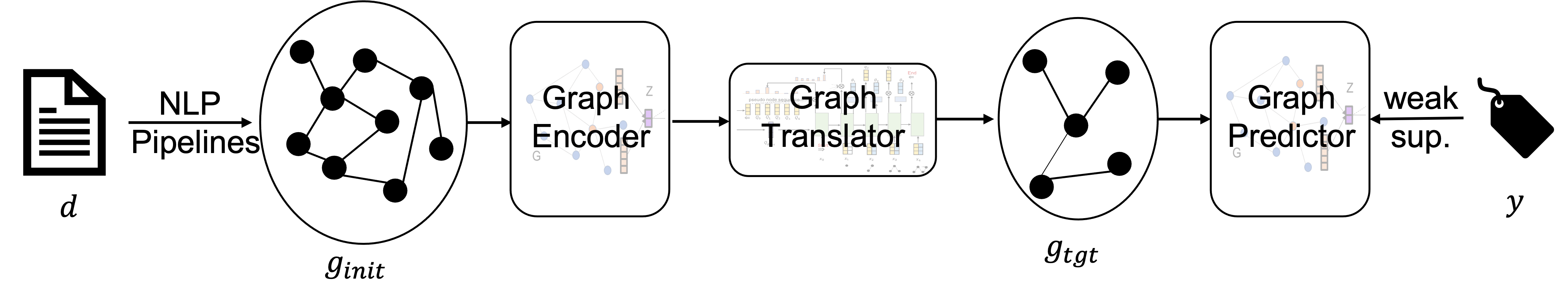}
    \caption{Overview of proposed \gtd framework.}
    \label{fig:framework}
\end{figure*}

\section{Related Work}
\label{sec:related-work}

\subsection{Automatic Concept Map Generation}
The concept map generation task is first introduced by \cite{DBLP:conf/emnlp/FalkeG17}, where the task definition and a benchmark dataset~\textit{EDUC} are proposed. In~\cite{DBLP:conf/emnlp/FalkeG17}, a corpus of 30 document clusters in which each contains around 40 source documents and 1 crowdsourcing summary concept map was provided. A keyphrase-based approach concept map generation approach was also proposed and evaluated in terms of precision, recall, and F1 of concept propositions (concept pairs). We do not include this approach as its follow-up study proposes a more advanced model. \textit{CMB-MDS}~\cite{falke-etal-2017-concept}, the extended model from the same research group, adapted the task definition, and then proposed an approach that utilized coreference resolution module to merge coreferent concepts and integer linear programming module to globally optimize the summary concept maps. 
Different from above mentioned studies, \dg~\cite{DBLP:conf/sigir/YangZWL020} did not rely on human-generated concept maps as training samples. Instead, the graph generation model in doc2graph was trained with weak supervision from downstream tasks.

\noindent\textbf{Side-by-side comparison between \dg and \gtd.} Both \dg and \gtd models are capable of generating concept maps without ground-truth training samples. However, \gtd is not a simple extension of \dg. As shown in Fig.~\ref{fig:intro_examples}, \gtd owns task-oriented, semantic-rich, concise, size-flexible, and label-efficient properties while \dg only owns two of them. The semantic-rich property is achieved by our proposed NLP pipeline constructed graph. The size-flexible property is achieved by the RNN-based edge decoder and the kernel-based length regularizer. The label-efficient property is achieved by the graph translator. Regarding the overall model designs, \gtd is based on the graph translation framework while \dg is based on generating graphs from scratch. In experiments, this work conducts additional extensive human evaluation, quantitative downstream application evaluation, labeling efficiency evaluation, flexibility evaluation, and comprehensive case studies, while \dg only conducts downstream application evaluation and simple case studies.

\subsection{Graphical Structures for Texts}
Graph structures have been extensively used in various NLP and text mining tasks, including keyword-based graph~\cite{mihalcea2004textrank,DBLP:journals/tkde/ShangLJRVH18} and parsing-based graph~\cite{DBLP:conf/acl/HopkinsJP18,he:20a,lu2021evaluation}. \textit{TextRank} \cite{mihalcea2004textrank} builds a word graph to represent the text, and connected words using a fixed-size sliding window. The word importance score is then calculated by a modified PageRank formula. \textit{TextRank} has been widely applied in keyword extraction and sentence extraction, and here we utilize it as a baseline for concept map generation. However, concepts are restricted as words and the number of concepts to keep cannot be inferred by the algorithm itself. \textit{AutoPhrase} \cite{DBLP:journals/tkde/ShangLJRVH18} is another popular method for keyword extraction and keyphrase mining tasks. Top-k high-quality phrases can be extracted as concepts and then a graph can be constructed using concepts co-occurring in the same sentences. Although~\textit{AutoPhrase} can generate both words and phrases as concepts, the drawback is that the size of the graphs has to be fixed.
In this work, we incorporate dependency parsing~\cite{lu2021evaluation} to provide the semantic-rich initial graph and propose a graph translation model to enable the size-flexible concept map generation.

\subsection{Graph Generation and Translation Methods} 
The focus of this paper, concept map generation, is a sub-task of the graph generation task. GraphVAE~\cite{DBLP:conf/icann/SimonovskyK18} and GraphRNN~\cite{you2018graphrnn} are two pioneering works that learn to a distribution model $p_{model}(\mathbb{G})$ over a set of observed graphs $\mathbb{G}=\{G_1,\dots,G_n\}$. Following this line, several flow-based models (GraphAF~\cite{GraphAF}, GraphDF~\cite{graphdf}, etc.) and diffusion models (GRAND~\cite{GRAND}, EDM~\cite{EDM}, etc.) are proposed for graph generation. However, these generative models all require a significant amount of ground-truth graph samples, while in concept map generation task these samples are often scarce. CondGen~\cite{DBLP:conf/cikm/YangGWSX019} explores generating novel graphs conditioned on unseen semantic labels by explicitly modeling the relationships between graph context and structures. \dg~\cite{DBLP:conf/sigir/YangZWL020} further tackles the graph generation under a weakly supervised manner. On the other hand, our proposed \gtd follows a graph translation (GT) paradigm to utilize the semantic-rich graph derived from NLP pipelines. VJTNN~\cite{jin2018VJTNN} proposes a VAE-based model for molecular graph translation. GT-GAN~\cite{guo2022GTGan} proposes a GAN-based model for directed weighted graph translation. Similarly, these models require high-quality paired datasets. SegTran~\cite{zhao2020segtran} is a semi-supervised autoencoder-based GT model that requires only a small fraction of training data. The basic idea of SegTran is to use a specific encoder/decoder for the source/target graph, and translate the graph in the latent domain. 
Our proposed \gtd further addresses the lack of training paired graph issue, and it is capable of generating concept maps from free-form texts using weak supervision from downstream tasks.

\section{Problems Statement}
\label{sec:problem-definition}

\header{Problem Definition}. We focus on the novel problem of weakly supervised concept map generation. It can be defined as follows: Given a text corpus $\mathcal{D}_l=(d_1,\dots,d_{i_l})$ with corresponding labels $\mathcal{Y}=(y_1,\dots,y_{i_l})$ of certain downstream text-related tasks, we aim at generating concept maps $g_i=\{\mathcal{C}_i, \mathcal{M}_i\}$ for each document $d_i\in D_{u}$ where $D_u$ is a set of unlabeled documents.
As can be seen from the definition, there are no ground-truth concept maps paired with the input text. Instead, weak or distant supervision from downstream tasks is provided. The downstream text-related tasks are very flexible, possibly ranging from document classification, retrieval, ranking, relation inference, \textit{etc}. 
The major output is concept maps $\mathcal{G}$ for all documents $\mathcal{D}=\mathcal{D}_l\cup \mathcal{D}_u$. A document $d\in \mathcal{D}$ is indeed a sequence of words, \textit{i.e.}, $d_i = (w_{i,1},\dots,w_{i,|d_i|})$. A concept map $g_i=\{\mathcal{C}_i, \mathcal{M}_i\}$ is an undirected graph which focuses on the concepts $\mathcal{C}_i$ and their interactions $\mathcal{M}_i$ in the span of $d_i$.
$\mathcal{C}_i=(c_{i,1},\dots,c_{i,|c_i|})$ is a set of $n$ concepts that can be words, phrases, or sentence fragments depending on the downstream tasks, and $\mathcal{M}_i \subseteq \mathbb{R}^{n \times n}$ indicates the interaction strength (\textit{i.e.}, edge weight) among concepts in $\mathcal{C}_i$.
Moreover, the auxiliary output is the predicted labels $\mathcal{\hat{Y}}$ for unlabeled documents $\mathcal{D}_u$.

We summarize the essential challenges of weakly supervised concept map generation as two folds in the following. 

\header{Challenge 1.} \textit{How to construct concept maps without training samples?} The concept map is a natural structure for representing interactions of concepts introduced in a document. However, despite its promising utility, the usage of concept maps so far is still limited. One critical reason is the lack of available training samples to drive the data-eager graph generative models. Therefore, models should be able to efficiently generate without training samples.

\header{Remarks 1.} Unsupervised concept map generation models have made important progress toward automatically generating large-scale concept maps from text data. Concept maps would be more helpful if generated concepts and interactions are relevant to downstream tasks. Considering the downstream task labels come with the text corpus, it is important that models can explicitly utilize these signals to guide the concept map generation process.

\header{Challenge 2.} \textit{How to guarantee the quality of generated concept maps?} Quality is another important property if concept maps aim to thrive in successful applications. Ideally, high-quality concept maps should concisely distill and represent the key information in the text. On the other hand, the downstream task performance can partially reflect the quality of generated concept maps. 

\header{Remarks 2.} Human evaluation should be regarded as the major examination criteria for the quality of generated concept maps, since no ground-truth concept maps are available in the datasets. Multiple metrics that cover different quality aspects are proposed and described in $\S$\ref{ssec:human-evaluation}. Moreover, automatic metrics $f(\mathcal{Y}_u, \hat{\mathcal{Y}}_u)$ (\eg, accuracy, MRR, \textit{etc}.) can be included as another indicator for quality evaluation.

\section{Proposed Approach} 
\label{sec:approach}

Fig. \ref{fig:framework} gives an overview of the proposed \textit{GT-D2G} (Graph Translation based Document-To-Graph) framework: A proper NLP pipeline is used to extract salient phrases from document $d$ and construct the initial semantic-rich concept map $g_{init}$. A Graph Encoder then encodes each node of $g_{init}$ into a node-level embedding $\bm{Q_i}$, and also represents the whole $g_{init}$ as a dense vector by aggregating all its node embeddings. A Graph Translator is responsible to identify the nodes needed to be kept in the target graph $g_{tgt}$ as well as proposing links among kept nodes iteratively. Once the nodes and links are generated, the target graph $g_{tgt}$ is fed into a Graph Predictor to produce a document label $\hat{y}$, which can be trained towards the ground-truth label $y$. The whole encoder-translator-predictor neural network is thus weakly supervised by the classification signal in an end-to-end fashion. In the following subsections, we expand with more technical details. 

\subsection{Enriching Concept Maps with Semantics}
\label{ssec:initial-map}

As we motivated before, one major drawback of doc2graph \cite{DBLP:conf/sigir/YangZWL020} is that single words are directly picked from the raw texts through a Pointer Network \cite{vinyals2015pointer} and considered as nodes in the final concept map. However, words purely picked by a simple Pointer Network can easily be of low-quality \cite{wang2019concept}. Moreover, phrases are often preferable to represent concepts, especially noun phrases as semantically complete concepts \cite{DBLP:journals/tkde/ShangLJRVH18}. 
For instance, extracting two nodes ``deep'', ``learning'' from a computer science paper is incomplete while ``deep learning'' as one concept node is semantically more meaningful and accurate. Some researchers propose to concatenate words that occur adjacently in the input document as extracted phrases to solve this issue, although potential heuristic post-processing is needed. In \textit{GT-D2G}, we aim to enrich concept maps with semantics by leveraging existing NLP pipelines~\cite{lu2021evaluation}. For simplicity and generalization concerns, we intentionally choose the most popular yet reliable NLP tools for initial concept map construction, which can be further extended according to application scenarios.

\vspace*{0.5\baselineskip} 
\noindent \textbf{Node Generation.} To avoid complicated pre-processing, we use multiple classic NLP tools in \textit{GT-D2G} to extract noun phrases, verb phrases, and adjectives as node candidates in the initial concept map. Sentence segmentation, pos-tagging, lemmatization, and constituency parsing are conducted for every document. Since constituency parsing detects sub-phrases of given sentences, we then first extract basic noun phrases from constituency parsing results. The basic noun phrases extraction algorithm is deterministic so that any noun phrase not containing other noun phrases is considered valid. After all basic noun phrases are identified, verb phrases and adjectives remaining in the text are extracted. Other discourse units such as adverbs and prepositions are discarded since they typically do not contain much knowledge or information. Due to the fact that multiple words can refer to the same concept, determinants such as ``a'', ``an'', ``the'' are removed from the node mentions, and words are replaced by their lemmas. Moreover, pronouns need to be merged into coreferent mentions to obtain a clean initial concept map. Thus, the coreference resolution technique is used to resolve all pronoun expressions in documents. We use the popular Stanford CoreNLP~\cite{manning2014stanford} for all steps mentioned above.

\vspace*{0.5\baselineskip} 
\noindent \textbf{Link Generation.} For links between extracted nodes, we follow the sliding window idea introduced in keyphrase extraction studies \cite{mihalcea2004textrank}. Nodes that occur within a fix-sized sliding window are connected to each other. Therefore, the initial concept maps are undirected graphs $g_{init}=\{C_{init}, M_{init}\}$. The link construction module is flexible in \textit{GT-D2G} so that any algorithms can be applied to construct weighted links or directed links. For instance, we can directly use the whole parsing tree or filter out certain types of relations for link generation. The graph ensemble process is trivial once nodes and links are extracted.

\subsection{Task Guided Graph Translation}
\label{ssec:graph-translation}

\begin{figure}[htbp!]
    \centering
    \includegraphics[width=\linewidth]{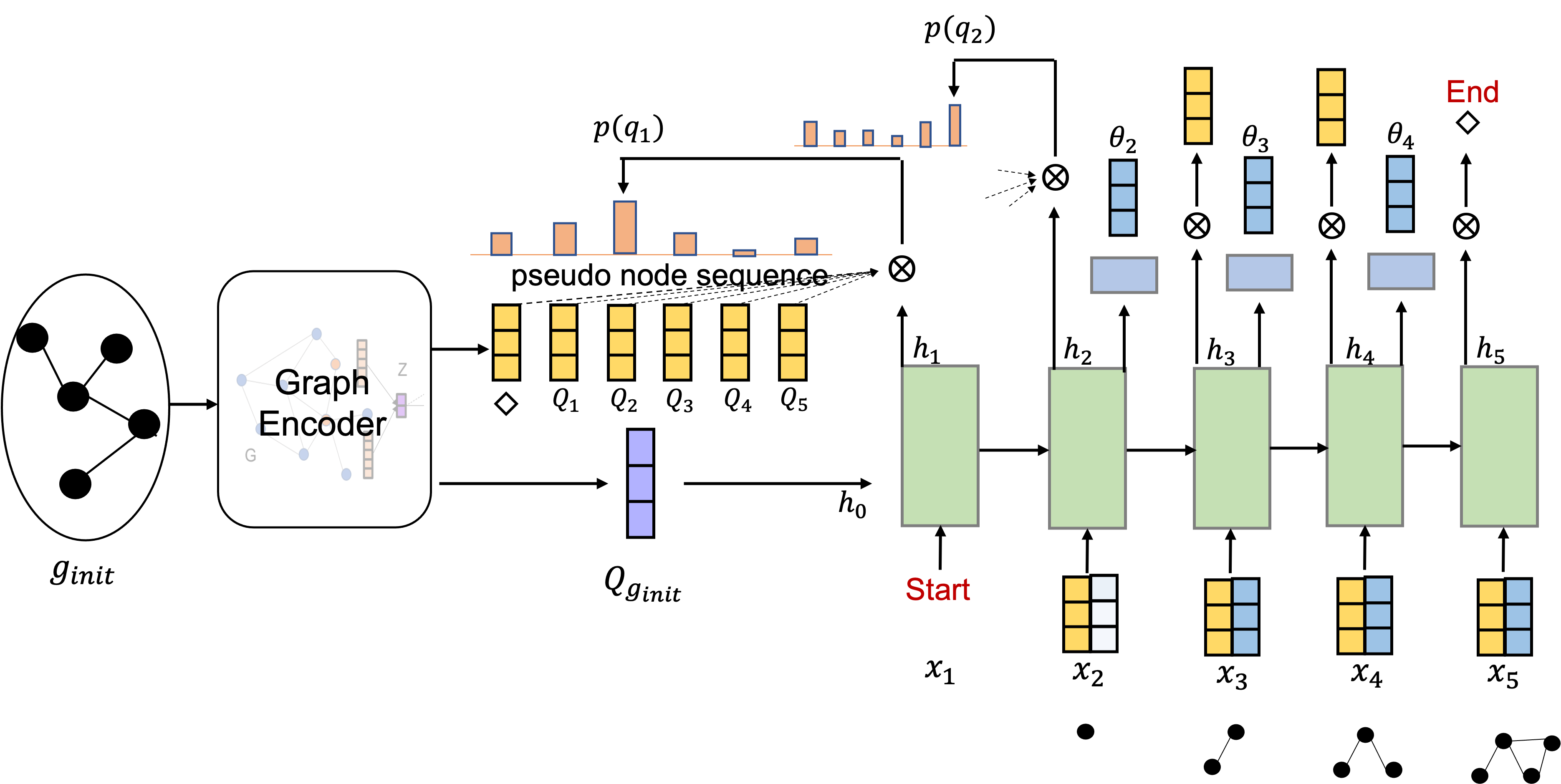}
    \caption{Graph Translator. Green rectangles denote RNN cells that take the previous time step chosen node $q_{t-1}$ and generated adjacency vector $\bm{\theta_{t-1}}$ as input. The RNN state vector $\bm{h_t}$ is updated at every time step, and is initialized by the graph level representation of initial graph $\bm{Q_{g_{init}}}$.}
    \label{fig:graph_translator}
\end{figure}

\vspace*{0.5\baselineskip}
\noindent \textbf{Graph Encoder.} Before graph translation, the model has to first learn to understand the initial graph. For this purpose, we adopt the recent successful graph representation learning model, \ie, Graph Convolutional Network (GCN) \cite{DBLP:conf/iclr/KipfW17} as our Encoder. The node embeddings $\bm{Q^{(k)}}$ are learned after the $k$-th layer of GCN by the following equation
\begin{equation}
    \bm{Q^{(k)}} = \ReLU(\Tilde{\bm{D}}^{-\frac{1}{2}}\Tilde{\bm{M}}\Tilde{\bm{D}}^{-\frac{1}{2}}\bm{Q^{(k-1)}}\bm{W_Q^{(k)}}),
\label{eq:gcn}
\end{equation}
where $\bm{W_Q^{(k)}}$ is learnable parameters in the $k$-th layer of GCN, $\Tilde{\bm{M}}\subseteq \mathbb{R}^{n \times n}$ is the adjacency matrix $\bm{M}_{init}$ with additional self-connections and $\Tilde{\bm{D}}_{ii}=\sum_j\Tilde{\bm{M}}_{ij}$ is the diagonal degree matrix. The input node embeddings $\bm{Q^{(0)}}$ are the concatenations of phrase embeddings, normalized frequency feature, and normalized location feature. The phrase embedding of each node is the average of pre-trained word embeddings (in practice, we use GloVe \cite{pennington2014glove}). The frequency feature and the location feature reflect the importance of the concept in the original text and are normalized by min-max scaling per graph. Besides the node-level embeddings, we also compute the graph-level embedding as $\bm{Q_{g_{init}}}=\frac{1}{n}\sum_{i=1}^{n}\bm{Q_i^k}$ to encode the global contextual information in the initial graph.

\vspace*{0.5\baselineskip}
\noindent \textbf{Graph Translator.} Our graph translator aims to choose the most informative nodes that are also beneficial to downstream tasks from the initial graph, while proposing links among the chosen nodes accordingly. In particular, the Graph Translator generates a sequence of nodes and their corresponding adjacency vectors based on the initial concept map $g_{init}$-- to be specific, its node-level embeddings $\bm{Q_i}\ (i \in [1,n])$ and graph-level embedding $\bm{Q_{g_{init}}}$ produced by the Graph Encoder. Since we expect to preserve the semantic rich and task-relevant concepts in the initial graph and only pick out a subset of nodes, we adopt the Pointer Network \cite{vinyals2015pointer} from keyword selection and novelly extend it into a graph version to generate a sequence of pointers for the selection of the most important nodes from the initial concept map. After each node is selected, we get inspiration from GraphRNN \cite{you2018graphrnn} to also generate its corresponding adjacency vector which contains links to previously selected nodes.  However, the original GraphRNN only works on the transductive learning setting when there is an actual graph as input to learning from. Therefore, we need to make several novel modifications to GraphRNN before seamlessly integrating it into our Graph Pointer Network (GPT) towards our novel setting of task-guided graph translation. 

\vspace*{0.25\baselineskip}
\noindent \textit{Graph Pointer Network.} Since the original Pointer Network \cite{vinyals2015pointer} works on sequential text data, we convert the non-sequential nodes in the initial concept map into a pseudo node sequence according to positions of node mentions in the source document, illustrated as the yellow bars in Fig. \ref{fig:graph_translator}. The order of pseudo node sequence is flexible and can be replaced with any other order for proper reasons (e.g., node degree order). Here we just follow the most intuitive way and do not observe significant performance differences when using other orders. 
In our GPT, we use a one-directional RNN decoder to model the process of translating a sequence of nodes and links from an initial graph, denoted as the green rectangles in Fig. \ref{fig:graph_translator}. In practice, we choose GRU \cite{DBLP:conf/emnlp/ChoMGBBSB14} as the implementation. In order to start the translation from the whole initial graph, the hidden state of the RNN decoder is initialized by $\bm{h_{0}}=\bm{Q_{g_{init}}}$, and the input of the first step is $\bm{x_1}=(0,\dots,0)^{\intercal}$. Therefore, the hidden state that encodes the ``graph translation state'' is updated by
\begin{equation} \label{eq:rnn}
    \bm{h_t}=\RNN(\bm{x_t},\bm{h_{t-1}}),
\end{equation}
where $\bm{h_{t-1}}$ denotes the hidden state from the last time step, and $\bm{x_t}$ denotes the input at the current time step. More specifically, we compute $\bm{x_t}$ as the representations of both nodes and links generated from the last time step, which can be denoted as 
\begin{equation}
\bm{x_t}=[\bm{q_{t-1}};\bm{\theta_{t-1}}],
\end{equation}
where $[\cdot;\cdot]$ denotes vector concatenation. $\bm{q_{t-1}=\bm{Q_i}}$ is the node embedding from the Graph Encoder of the last selected node $i$, and we defer the explanation towards adjacency vector $\bm{\theta_{t-1}}$ to the later part of this subsection.

\vspace*{0.25\baselineskip}
\noindent \textit{Deeply coupled node and link generation.} Once we obtain the RNN decoder hidden state $\bm{h_t}$, the node selection process can be described by the following equations

\begin{align} 
    \label{eq:node_select1}
    &e_{i,t} = \bm{v^{\intercal}}\tanh(\bm{W}[\bm{Q_i};\bm{h_t}]),\\
    \label{eq:node_select2}
    &\mathsf{p}_{t,i}=\mathsf{p}(\bm{q_t}=\bm{Q_i}) = \frac{\exp(e_{i,t})}{\sum_{j=1}^{n}\exp(e_{j,t})},
\end{align} 
where $\bm{v}\subseteq \mathbb{R}^{d_e}$ and $\bm{W}\subseteq \mathbb{R}^{d_h\times d_e}$ are learnable parameters for calculating the unnormalized node selection score $e_{\cdot,t}$ at time step $t$ for every node in initial graph node set $C_{init}$. 
Our GPT then selects the $i$-th node with the maximum score by 
\begin{equation} \label{eq:node_select3}
    i=\underset{i}{\argmax} (\mathsf{p}_{t,i}),
\end{equation}
adds the selected node into the translated target graph and feeds $\bm{q_t}=\bm{Q_i}$ into the RNN decoder at the next time step. To improve the semantic completeness of selected concept nodes, we also adapt the coverage loss in \cite{tu2016modeling}, by maintaining a coverage vector $\bm{c_t}=\sum_{t'=0}^{t-1}\mathsf{p}(\bm{q_{t'}})$ that accumulates the generated attention so far, while adding the following loss to enforce the model to pay more attention to nodes not covered yet:

\begin{equation}
    L_{cov} = \sum_{d_i\in \mathcal{D}}\sum_{t_j\in d_i}\min(\mathsf{p}(\bm{q_{t'}}), \bm{c_{t_j}}).
\end{equation}

To deeply couple the generation process of nodes and links so that the target graph (\ie, final concept map) is meaningful, we get inspired by the recent deep graph generation model of GraphRNN \cite{you2018graphrnn}. Specifically, in our GPT, at each time step, after a new node is generated, we immediately generate its associated adjacency vector regarding all links between it and all previously generated nodes, as denoted by smaller blue rectangles in Fig. \ref{fig:graph_translator} and described in the following equation

\begin{equation} \label{eq:edge_propose}
    \begin{array}{l}
         \bm{\theta_t}=f_{out}(\bm{h_t}),
    \end{array}
\end{equation}
where $\bm{\theta_{t}}$ is the length $t-1$ adjacency vector for the chosen node at time step $t$ that is output by $f_{out}$. Based on slightly different goals for link generation, we design two variants of $f_{out}$: the \textit{path} variant and the \textit{neigh} variant. The former models the adjacency vector generation as generating a path connecting some previously picked nodes to the currently picked one, focusing on the higher-order sequential information among concepts. Hence, $f_{out}^{path}$ is implemented as another RNN that connects to the hidden state of the RNN decoder. On the other hand, the \textit{neigh} variant interprets the generation problem as generating all possible neighbors of the currently picked node from all previously picked nodes, focusing on the first-order neighborhood structures of concepts.  Therefore, $f_{out}^{neigh}$ is implemented as a multi-layer perceptron (MLP) with non-linear activation. The weights of $f_{out}$ are shared across all time steps to reduce the number of parameters and alleviate overfitting. In our experiments, we find the \textit{neigh} variant to be preferable over the \textit{path} variant, which can be intuitively attributed to the fact that structural information is more important than sequential information among concepts. 

\vspace*{0.5\baselineskip}
\noindent \textbf{Graph Predictor.} After generating a sequence of nodes $q_1,\dots,q_T$ and adjacency vectors $\theta_1,\dots,\theta_T$, we assemble the target graph as 
\begin{equation} \label{eq:graph_assemble}
    g_{tgt}=\{C_{tgt},M_{tgt}\}=\{(q_1,\dots,q_T), (\theta_1,\dots,\theta_T)\}.
\end{equation}
For the downstream graph-level prediction, we adopt a \textit{Graph Isomorphism Network} (GIN)~\cite{DBLP:conf/iclr/XuHLJ19} due to GIN's superior discriminative power to capture different graph structures. More specific, we adopt the sum operator as the neighborhood aggregation function, and an MLP as the center node and neighbor nodes combination function:
\begin{equation} \label{eq:GIN}
    \bm{q^{(k)}}=\ReLU((\bm{M_{tgt}}+(1+\epsilon^{(k)})\bm{I})\bm{q^{(k-1)}}\bm{W_q^{(k)}}),
\end{equation}
where $\bm{M_{tgt}}\subseteq \mathbb{R}^{T\times T}$ is the adjacency matrix of translated concept map, $\bm{I}\subseteq \mathbb{R}^{T\times T}$ is a identity matrix (\ie, self connection), and $\epsilon^{(k)}, \bm{W_q^{(k)}}$ are learnable parameters for GIN's k-th layer.
Furthermore, the graph label (\ie, document category in our case) $\hat{y}$ is obtained by an additional two-layer MLP on the graph representation: 
\begin{equation} \label{eq:graph_pred}
    \hat{y}=\MLP(\text{concat}(\text{sum}(\bm{q^{(k)})} | k=1,\dots,K)),
\end{equation}
where the graph representation is achieved by summing all node embeddings from the same layer, and then concatenating summed embeddings across all layers.

\subsection{Training Techniques} \label{ssec:train_tech}
The whole model is trained in a weakly supervised end-to-end fashion, by computing the cross-entropy loss for the downstream task-- document classification as we focus on in this work, and the coverage loss for the node selection in our GPT. Specifically, we have
\begin{align}
    L_{cls} &= -\sum_{d_i\in \mathcal{D}} \mathsf{p}(\hat{y_i})\log \mathsf{p}(y_i), \\
    L &= L_{cls} + \lambda * L_{cov},
\end{align}
where $\lambda$ is a tunable hyper-parameter. 

One technical challenge exists for the node selection operation that selects the node with maximum pointer attention $i=\underset{i}{\argmax} (\mathsf{p}(\bm{q_t}=\bm{Q_i}))$ during the graph translation process in GPT. Firstly, the max value selection operation implemented as \textit{argmax} is non-differentiable, thus leading to the lost gradient after node selection. Secondly, \textit{argmax} is a deterministic sampling operator, thus making the GPT loses exploration ability. The exploration ability or stochastic sampling is important during the early training stages of GPT, because the predicted probability to select a node is not very reliable at that time. Inspired by the re-parameterization tricks for categorical variables sampling~\cite{DBLP:journals/corr/BengioLC13,DBLP:conf/iclr/MaddisonMT17,DBLP:conf/iclr/JangGP17}, we adopt a hard-version \textit{Gumbel-Softmax} to sample one-hot vectors from the predicted probabilities, so that the node selection process in GPT is differentiable and stochastic. The sampled probability $P_{t,i}$ to choose node $i$ at time step $t$ then becomes: 
\begin{equation}
    P_{t,i}=\softmax(\log(\mathsf{p}_{t,i}) + G_i, \tau), 
\end{equation}
where $\mathsf{p}_{t,i}$ is the predicted probability as defined in Eq~\eqref{eq:node_select2}, $G_i\sim \text{Gumbel}(0,1)$ is the $i$-th random variable sampled from the Gumbel distribution, and $\tau$ is the temperature parameter for \textit{softmax}. We set a relative large temperature to enforce $P_t=(P_{t,1}, P_{t,2}, \dots, P_{t,n})$ has the one-hot vector shape. During training, we use $P_t \cdot Q$ to represent selecting one particular node for gradient backpropagation.

Moreover, to generate concept maps of flexible sizes, we incorporate the special \textit{``EOS''} node at the first position of pseudo node sequence, denoted as ``$\Diamond$'' in Fig. \ref{fig:graph_translator}. The end of an output node sequence is determined when the \textit{``EOS''} is predicted. For the completeness of concept maps, we penalize node sequences that are too short, which can be implemented by applying a penalty to \textit{``EOS''} node predicted at every time step as follows
\begin{equation}
    L_{len} = \sum_{d_i\in \mathcal{D}}\sum_{t_j\in d_i} \Penalty(t_j)\cdot \mathsf{p}(\bm{q_{t_j}}=``EOS").
\end{equation}
The function $\Penalty(t) > 0$ defines a penalty curve depending on the current time step $t$. In our implementation, we choose the RBF kernel function $\Phi (t,t')=exp(-\frac{\left\|t-t'\right\|^2}{2\sigma^2})$ for the penalty curve \cite{chen1991orthogonal}. Therefore, the overall 
loss function for \textit{GT-D2G} is:

\begin{equation} \label{eq:overall_loss}
    L = L_{cls} + \lambda_1 * L_{cov} + \lambda_2 * L_{len}.
\end{equation}

To sum up, our whole framework is trained in an end-to-end fashion, while Graph Encoder, Graph Translator, and Graph Predictor are guided by the downstream task with the goal of reducing classification loss. In this way, each module is jointly learned and enhanced. Moreover, the translation process is regularized by the coverage loss and graph size loss, aiming to produce high-quality concept maps depending on the input documents' characteristics.

\begin{algorithm}[h!]
\caption{\gtd Training Algorithm} \label{algorithm}
\KwData{initial concept map $g_{init}=\{\mathcal{C}_{init}, \mathcal{M}_{init}\}$, input node embeddings $\{Q^{0}_{v},\forall v \in \mathcal{C}_{init}\}$, ground truth graph label $y$}
\KwResult{translated concept map $g_{tgt}=\{\mathcal{C}_{tgt},\mathcal{M}_{tgt}\}$, predicted graph label $\hat{y}$}
\BlankLine
\emph{Initialize \gtd parameters}\;
\While{not converge}{
  \tcc{Obtain graph representation of $g_{init}$}\
  Update node embedding $\bm{Q^{(k)}}=\text{GCN}_{\text{Enc}}(\bm{Q^{(0)}};k)$ by Eq.~\ref{eq:gcn}\;
  Update graph embedding $\bm{Q_{g_{init}}}= \text{pooling}(\{\bm{Q_v^{(k)}},\forall v \in \mathcal{C}_{init}\})$\;
  \tcc{Translate $g_{init}$ into $g_{tgt}$ step-by-step}\
  \While{not generate \textit{``EOS''} node}{
     Prepare Graph Translator (RNN) input $(\bm{x_t}, \bm{h_{t-1}})$ by initilization or previous step results\;
     Update hidden state of Graph Translator $\bm{h_t}$ by Eq.~\eqref{eq:rnn}\;
     Generate node $q_t$ by Eq.~\eqref{eq:node_select1},~\eqref{eq:node_select2},~\eqref{eq:node_select3}\;
     Generate adjacency vector $\theta_t$ by Eq~\eqref{eq:edge_propose}\;
  }
  \tcc{Predict graph label}\
  Assemble the translated concept map $g_{tgt}$ by Eq.~\eqref{eq:graph_assemble}\;
  Predict the graph label $\hat{y}$ by Eq.~\ref{eq:GIN},~\ref{eq:graph_pred}\;
  \tcc{Backpropagate the weak supervision}\
  Compute the overall loss $L$ by Eq.~\eqref{eq:overall_loss}\;
  Update model parameters with the gradients of $L$.
}
\end{algorithm}

\subsection{Complexity Analysis}
To analyze the computational efficiency of the proposed model, we present the \gtd training algorithm for one input initial concept map (one input document). The actual implementation is based on mini-batch training, and is publicly available\footnote{\label{fn:gtd}\gtd: \url{https://github.com/lujiaying/GT-doc2graph}}.
For obtaining graph representation of the initial concept map (L3-L4), the time complexity is $\mathcal{O}(Knd^2+Kmd)$, where $K$ is the number of GCN encoder layers, $d$ is the embedding dimensions (128 in all layers), $n$ is the number of nodes in $g_{init}$ (tens of nodes in our experiments), $m$ is the number of edges in $g_{init}$ (\eg, close to one hundred edges in our experiments). The time complexity can be further simplified into $\mathcal{O}(Knd^2)$ since $nd\gg m$.
For graph translation (L5-L9), the time complexity is $\mathcal{O}(TKnd^2)$, where $T$ is the size of the translated concept map, $K$ is reused to represent the number of RNN decoder layers (\eg, we set both GCN encoder, RNN decoder and GIN classifier layer sizes as 2), $d$ is reused to represent the RNN embedding dimensions (\eg, we set the hidden dimension to 128 for all modules).
For the graph label prediction(L10-L11), the time complexity is $\mathcal{O}(KTd^2)$ which is similar to GCN encoder analysis.
Therefore, the overall time complexity for proposed \gtd is $\mathcal{O}(Knd^2+TKnd^2+KTd^2)=\mathcal{O}(TKnd^2)$.

It is worth noting that the construction of initial concept maps is quite efficient, as the toolkit we employed (\eg, JVM-based Stanford CoreNLP~\cite{manning2014stanford}) mainly utilize pre-trained models or rule-based annotators for the NLP pipelines. Moreover, \dg's time complexity is $\mathcal{O}(TK\|\mathcal{D}\|d^2)$, where $\|\mathcal{D}\|$ denotes the number of words of input document. \gtd is more efficient than \dg, due to the fact that $\|\mathcal{D}\| \ge n$ in most cases. However, the advantage of \dg is that it does not require NLP pipelines to derive the initial concept maps.
\section{Experiments}
\label{sec:experiments}

In this section, we evaluate our proposed \textit{GT-D2G} framework focusing on the following four research questions:

\noindent \textbf{\textit{RQ1}}: How is the quality of \textit{GT-D2G} generated graphs?

\noindent \textbf{\textit{RQ2}}: How do \textit{GT-D2G} and its variants perform in comparison to other document classification methods?

\noindent \textbf{\textit{RQ3}}: Is \textit{GT-D2G} label efficient?

\noindent \textbf{\textit{RQ4}}: Can \textit{GT-D2G} generate flexible sizes of concept maps?

\subsection{Experiment Settings}
\label{ssec:experiment-settings}

\noindent \textbf{Datasets}. Our experiments are conducted on three real-world text corpora~\cite{DBLP:conf/sigir/YangZWL020}: \textit{NYT}, \textit{AMiner}, and \textit{Yelp}.
Different from ~\cite{DBLP:conf/sigir/YangZWL020}, for the \textit{Yelp} dataset, we re-grouped the 1-5 star reviews into negative, neutral and positive ratings.
The statistics of the three datasets are listed in Table~\ref{tab:stat}. 
For standard document classification, we follow the setting in~\cite{DBLP:conf/sigir/YangZWL020} to randomly split the labeled documents into 80\% for training, 10\% for validation, and 10\% for testing. We choose accuracy as the metric for document classification tasks. 
To get a stable result, we run each model three times and report the mean $\pm$ standard deviation. 

\begin{table}[ht!]
\caption{Statistics of three datasets.}
\resizebox{\linewidth}{!}{%
\begin{tabular}{c|ccc|ccc}
\hline
\multirow{2}{*}{Dataset} & \multirow{2}{*}{\#doc} & \multirow{2}{*}{\#word} & \multirow{2}{*}{\#category} & \multicolumn{3}{c}{Init Concept Map} \\
        &        &       &    & \#node & \#edge & \#degree \\ \hline\hline
NYT & 13,081 & 88.64 & 5 & 34     & 84     & 4.9      \\
Aminer  & 21,688 & 87.27 & 6  & 34     & 81     & 4.8      \\
Yelp    & 25,357 & 71.59 & 3  & 28     & 76     & 5.4      \\ \hline
\end{tabular}%
}
\label{tab:stat}
\end{table}

\noindent \textbf{Compared Methods}. We compare \textit{GT-D2G} with two sets of baselines described as follows: 

\noindent \textit{Graph-Based Methods} as major competitors.
    \begin{itemize}[labelindent=0pt]
        \item \textbf{\textit{AutoPhrase}} \cite{DBLP:journals/tkde/ShangLJRVH18}: This is a Pos-Guided Phrasal Segmentation model for phrase mining. We use the top-n highest quality phrases mined from input text as concepts and connect concepts in same sentence. The edge weights is computed as $w_{ij}=1-e^{-c_{ij}}$, where $c_{ij}$ denotes sentence-level co-occurring times of concept i and j.
        \item \textbf{\textit{TextRank}} \cite{mihalcea2004textrank}: A word co-occurrence graph is first constructed using a sliding window that connects any two words within the window. We use words with top-n maximum PageRank values as concepts. The edge weights are computed in the same way as \textit{AutoPhrase}. 
        \item \textbf{\textit{CMB-MDS}} \cite{falke-etal-2017-concept}: We use its pipeline to construct concept map and filter out concepts with low importance scores to keep top-n concepts. The edge weights are set to 1 according to the \textit{CMB-MDS} implementation.
        \item \textbf{\textit{doc2graph}} \cite{DBLP:conf/sigir/YangZWL020}: \textit{doc2graph} is a neural concept map generation model that is capable of generating concept maps through distant document classification supervision. We follow their implementation to pre-define graph size as n. 
    \end{itemize}
    
\noindent \textit{Text-Based Methods} as performance benchmarks.
    \begin{itemize}[labelindent=0pt]
        \item \textbf{\textit{Bi-LSTM}} \cite{graves-schmidhuber:2005}: \textit{Bi-LSTM} is a commonly used RNN model in text classification that learns the long-term dependencies in the document. We train \textit{Bi-LSTM} on the training set using the output from last time-step to predict document categories.
        \item \textbf{\textit{BERT-base}} \cite{devlin-etal:2019}: \textit{BERT} has achieved excellent performance on a wide range of NLP tasks as a state-of-the-art language model. In our experiment, We fine-tune the pre-trained \textit{BERT-base} model on the classification task.  
    \end{itemize}

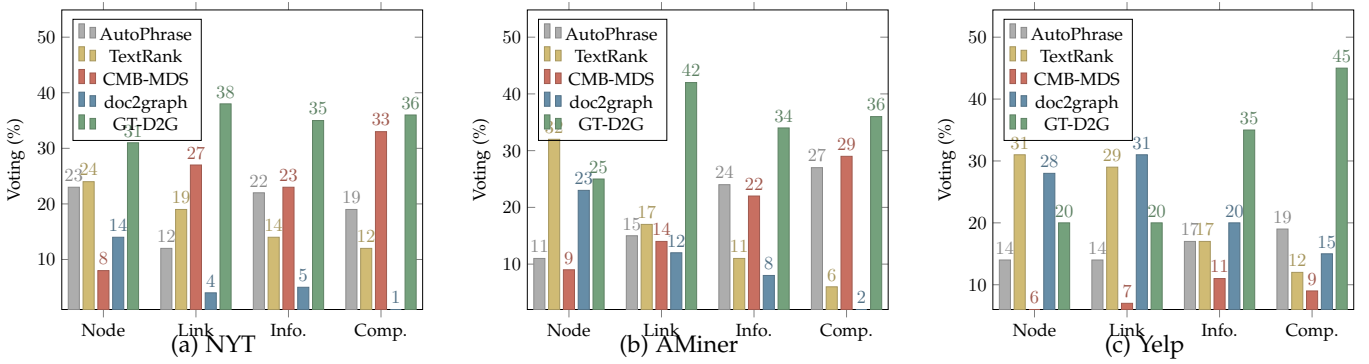
\begin{figure*}[htbp!]
\centering
\begin{subfigure}[b]{0.32\textwidth}
\centering
\begin{tikzpicture}[scale=0.7]
\begin{axis}[
    ybar,
    ymax=50,
    enlarge y limits={upper=0},
    enlarge x limits=0.15,
    legend pos=north west,
    legend style={fill=none},
    ylabel={Voting (\%)},
    symbolic x coords={Node, Link, Info., Comp.},
    xtick=data,
    bar width=6pt,
    nodes near coords,
    nodes near coords align={vertical},
    % x tick label style={rotate=45,anchor=east},
]
\addplot+ [yrb0!80!black, fill=yrb0] coordinates {(Node,23) (Link,12) (Info.,22) (Comp.,19)};
\addplot+ [yrb1!80!black, fill=yrb1]  coordinates {(Node,24) (Link,19) (Info.,14) (Comp.,12)};
\addplot+ [yrb2!80!black, fill=yrb2] coordinates {(Node,8) (Link,27) (Info.,23) (Comp.,33)};
\addplot+ [yrb3!80!black, fill=yrb3] coordinates{(Node,14) (Link,4) (Info.,5) (Comp.,1)};
\addplot+ [yrb4!80!black, fill=yrb4]  coordinates {(Node,31) (Link,38) (Info.,35) (Comp.,36)};
    \legend{AutoPhrase,TextRank,CMB-MDS,doc2graph,GT-D2G}
\end{axis}
\end{tikzpicture}
\vspace{-0.8cm}
\caption{NYT}
\label{tab:nyt_human_evaluate}
\end{subfigure}
\hspace{\fill}
\begin{subfigure}[b]{0.32\textwidth}
\centering
\begin{tikzpicture}[scale=0.7]
\begin{axis}[
    ybar,
    ymax=50,
    enlarge y limits={upper=0},
    enlarge x limits=0.15,
    % legend style={at={(0.5,-0.2)},anchor=north,legend columns=-1},
    % legend cell align=right,
    % legend pos=outer north east,
    legend pos=north west,
    legend style={fill=none},
    ylabel={Voting (\%)},
    symbolic x coords={Node, Link, Info., Comp.},
    xtick=data,
    bar width=6pt,
    nodes near coords,
    nodes near coords align={vertical},
    % x tick label style={rotate=45,anchor=east},
]
\addplot+ [yrb0!80!black, fill=yrb0]  coordinates {(Node,11) (Link,15) (Info.,24) (Comp.,27)};
\addplot+ [yrb1!80!black, fill=yrb1]  coordinates {(Node,32) (Link,17) (Info.,11) (Comp.,6)};
\addplot+ [yrb2!80!black, fill=yrb2]  coordinates {(Node,9) (Link,14) (Info.,22) (Comp.,29)};
\addplot+ [yrb3!80!black, fill=yrb3]  coordinates {(Node,23) (Link,12) (Info.,8) (Comp.,2)};
\addplot+ [yrb4!80!black, fill=yrb4] coordinates {(Node,25) (Link,42) (Info.,34) (Comp.,36)};
    \legend{AutoPhrase,TextRank,CMB-MDS,doc2graph,GT-D2G}
\end{axis}
\end{tikzpicture}
\vspace{-0.8cm}
\caption{AMiner}
\label{tab:aminer_human_evaluate}
\end{subfigure}
\hspace{\fill}
\begin{subfigure}[b]{0.32\textwidth}
\centering
\begin{tikzpicture}[scale=0.7]
\begin{axis}[
    ybar,
    ymax=50,
    enlarge y limits={upper=0},
    enlarge x limits=0.15,
    legend pos=north west,
    legend style={fill=none},
    ylabel={Voting (\%)},
    symbolic x coords={Node, Link, Info., Comp.},
    xtick=data,
    bar width=6pt,
    nodes near coords,
    nodes near coords align={vertical},
    % x tick label style={rotate=45,anchor=east},
    every axis plot/.append style={fill},
    cycle list/Paired,
]
\addplot+ [yrb0!80!black, fill=yrb0] coordinates {(Node,14) (Link,14) (Info.,17) (Comp.,19)};
\addplot+ [yrb1!80!black, fill=yrb1] coordinates {(Node,31) (Link,29) (Info.,17) (Comp.,12)};
\addplot+ [yrb2!80!black, fill=yrb2] coordinates {(Node,6) (Link,7) (Info.,11) (Comp.,9)};
\addplot+ [yrb3!80!black, fill=yrb3] coordinates {(Node,28) (Link,31) (Info.,20) (Comp.,15)};
\addplot+ [yrb4!80!black, fill=yrb4] coordinates {(Node,20) (Link,20) (Info.,35) (Comp.,45)};
    \legend{AutoPhrase,TextRank,CMB-MDS,doc2graph,GT-D2G}
\end{axis}
\end{tikzpicture}
\vspace{-0.8cm}
\caption{Yelp}
\label{tab:yelp_human_evaluate}
\end{subfigure}
\caption{Human evaluation results on (a) NYT, (b)AMiner, (c)Yelp based on four proposed metrics.}
\label{fig:human_evaluate}
\end{figure*}

\noindent \textbf{Implementation Details}. We implement \textit{GT-D2G} using Pytorch \cite{NEURIPS2019_9015} and DGL \cite{wang2019dgl}, with code publicly available\textsuperscript{\ref{fn:gtd}}. 
Implementations of the compared baselines are either from open-source project (\textit{BERT}\footnote{\textit{BERT}: \url{https://github.com/huggingface/transformers}}) or the original authors (\textit{Bi-LSTM}/ \textit{AutoPharse}/ \textit{TextRank}/ \textit{CMB-MDS}/ \dg\footnote{\dg: \url{https://github.com/JieyuZ2/doc2graph}}). We optimize \textit{GT-D2G} through the Adam optimizer with learning rate to $3e-4$ and max epoch to $500$. 
The temperature parameter $\tau$ for Gumbel-softmax starts from a big number (\textit{e.g.} 3 or 5) and then anneals along with training epochs to encourage exploration on the later stage. To get a higher accuracy, we set batch size to 64 for training. The hidden layer dimension of GCN, RNN and MLP are set to $128$, and the number of GNN layers in all GCN, GIN models are $2$. For RBF kernel function used to penalize overlength node sequence, $\sigma$ and $t_{prime}$ are set to $4$ and $0$, respectively. We choose GRU for RNN used in generating nodes and edges for simplicity sake. All other hyper-parameters are tuned separately on the validation set.

\subsection{Human Evaluation (\textit{RQ1})}
\label{ssec:human-evaluation}

\begin{table}[htbp!]
%\vspace*{\baselineskip} 
\caption{Correlation coefficients among the five peer annotators with manual responsiveness scores on a total of 300 documents of NYT, AMiner, Yelp (100 each).}
\centering
\begin{tabular}{c|cccc}
\toprule
Peer Scoring & Node & Link & Info. & Comp. \\ \hline
NYT          & 0.50 & 0.89 & 0.57  & 0.67  \\
AMiner       & 0.76 & 0.80 & 0.75  & 0.93  \\
Yelp         & 0.73 & 0.79 & 0.70  & 0.92  \\ 
\bottomrule
\end{tabular}
\label{tab:human-evaluation}
\end{table}

Human evaluation is critical to answer \textit{RQ1}, \textit{i.e.} evaluating the quality of generated concept maps, since there are no ground-truth concept maps on the three document classification datasets. Five expert annotators are hired to evaluate graphs generated from the text data by five methods: \textit{AutoPhrase}, \textit{TextRank}, \textit{CMB-MDS}, \textit{doc2graph}, and \textit{GT-D2G}. More specifically, on each dataset, we randomly sample 100 document with associated graphs of each method. or each document, annotators are asked to rank the five concept maps in terms of four metrics:

\noindent \textbf{Node}: regardless of downstream tasks, whether nodes are semantic complete, in proper length and not redundant. \\
\noindent \textbf{Link}: whether links between nodes are consistent with the text and make sense. \\
\noindent \textbf{Informativeness}: whether the generated graph is helpful for the downstream task. \\
\noindent \textbf{Completeness}: whether the generated graph covers the most salient information of the original text from different aspects.

Correlation Coefficient is a widely used indicator to estimate the inter-annotator agreement (ITA). However, we observe that explicitly annotating the rank among all five concept maps leads to low inter-annotator agreement. Therefore, we allow annotators to pick $k$ ($k\leq 3$) graphs for each metric as top graphs, as long as they think these k graphs are of the same best quality. That means, if an annotator thinks two graphs by \textit{doc2graph} and \textit{GT-D2G} are competitive in Informativeness, she can mark both two as top graphs without distinguishing which is the best. The top max-k graph annotation guideline gives high Correlation Coefficient scores, as can be seen in
Table \ref{tab:human-evaluation}.

The human evaluation results are shown in Fig.~\ref{fig:human_evaluate}. The value on y-axis indicates the percentage of the data that the annotator think the method performs best under the corresponding metric. For the metrics of \textit{Informativeness} and \textit{Completeness}, annotators reached a high degree of consistency that our approach \textit{GT-D2G} outperforms other baseline methods significantly. Moreover, \textit{GT-D2G} performs best on \textit{NYT} for \textit{Node} metrics and \textit{NYT} and \textit{AMiner} for \textit{Link} metrics.

\begin{figure*}[!]
\centering
\includegraphics[width=\linewidth]{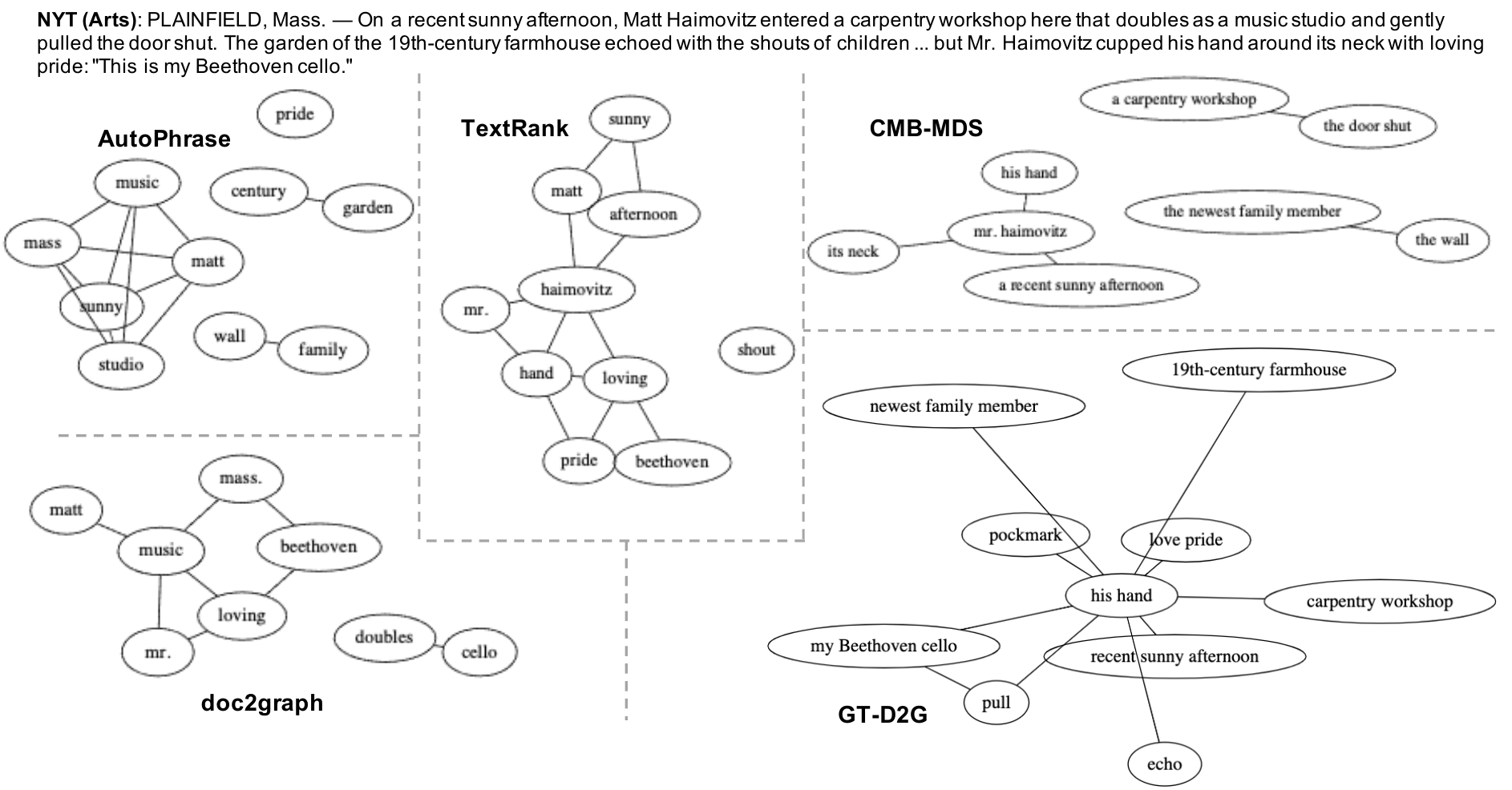}
\caption{Concept maps generated by various models for case studies.}
\label{fig:case_study}
\end{figure*}

\begin{figure*}[p!]
\centering
  \begin{subfigure}[b]{\linewidth}
     \centering
     \includegraphics[width=\linewidth]{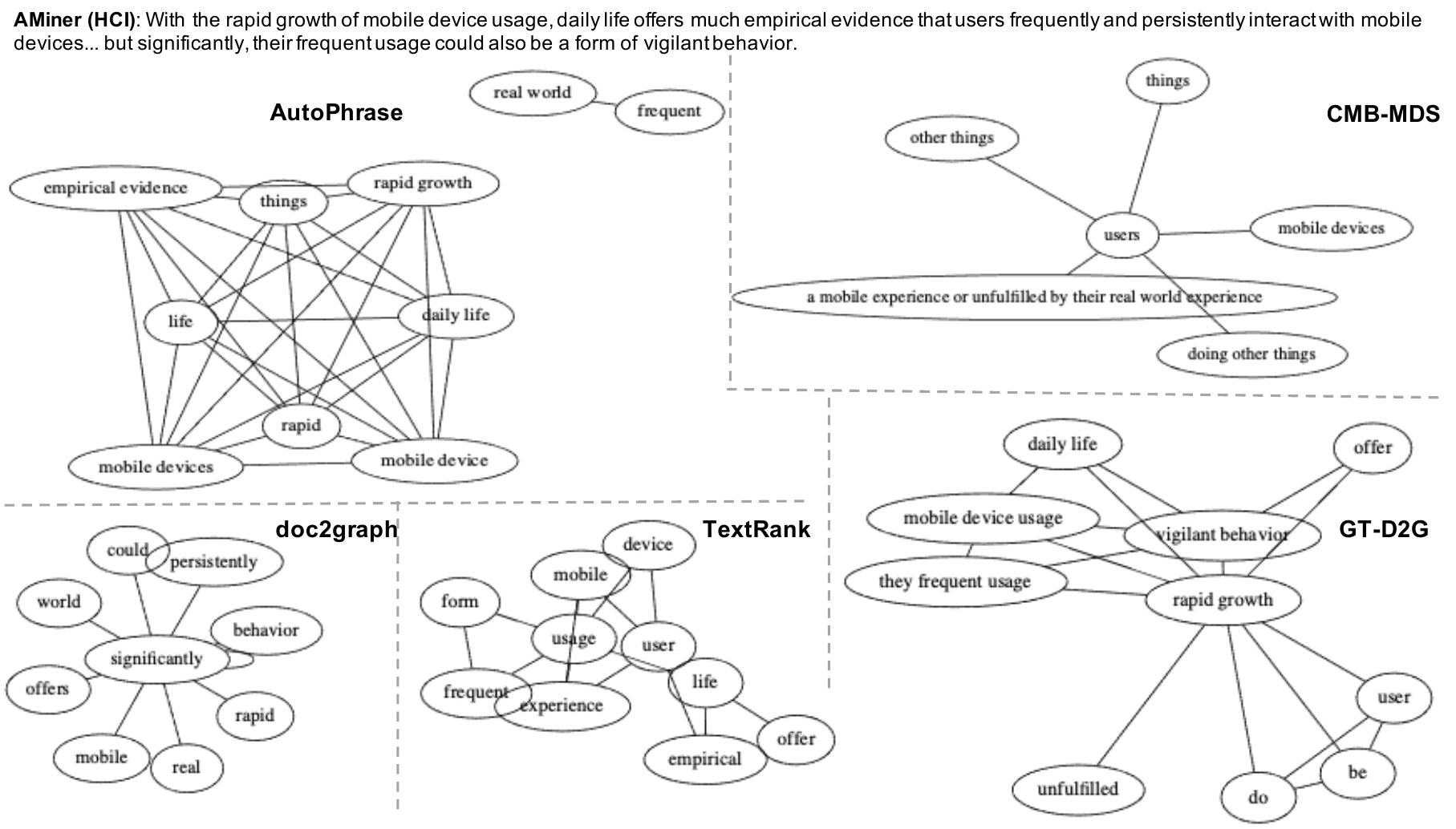}
  \end{subfigure}
  \par \bigskip \bigskip
  \begin{subfigure}[b]{\linewidth}
     \centering
     \includegraphics[width=\linewidth]{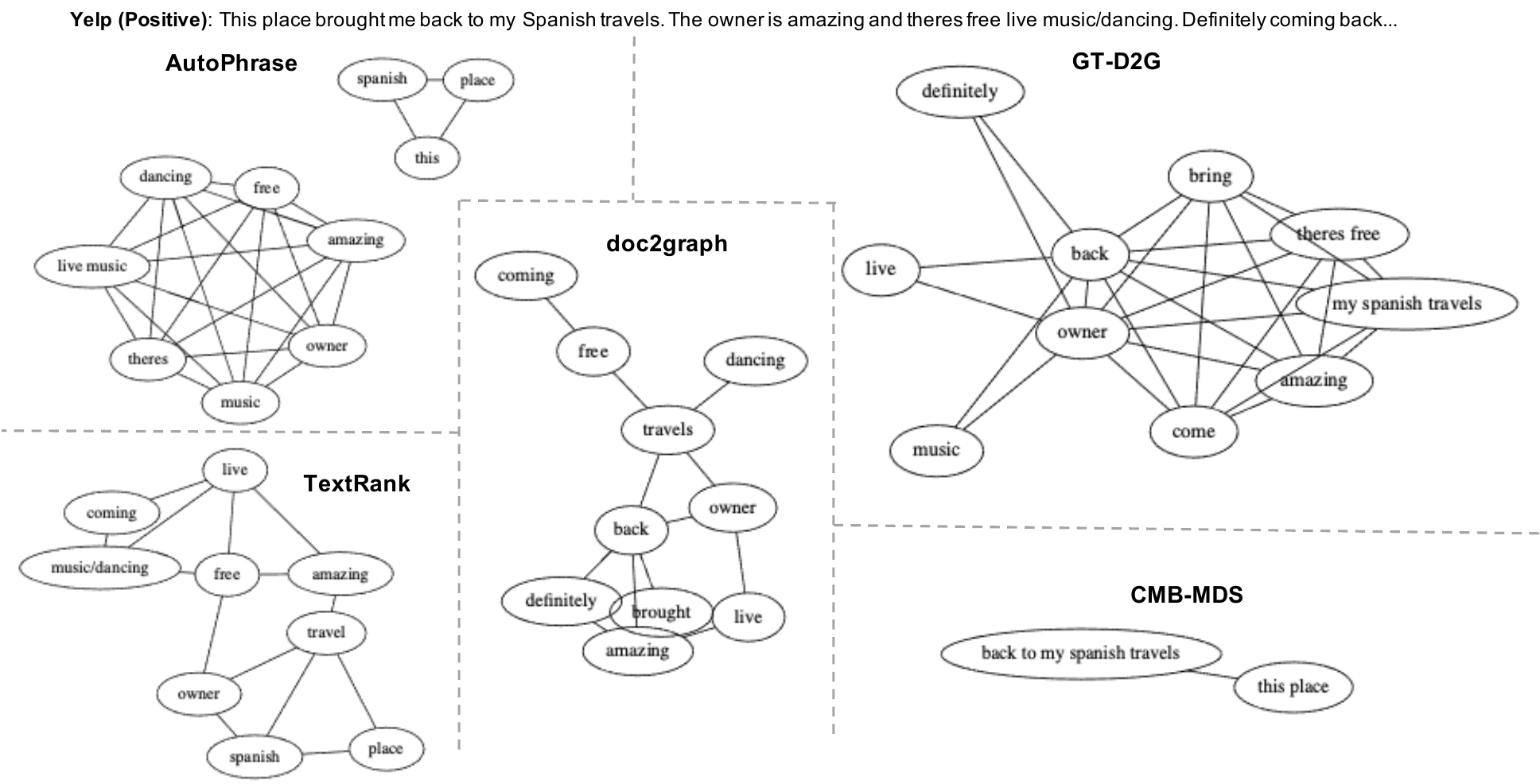}
  \end{subfigure}
  \caption{Concept maps generated by various models for case studies (\textit{cont.}).}
  \label{fig:case_study_cond}
\end{figure*}

\noindent
\textbf{Case Studies}. The concept maps constructed by five methods are shown in Fig.~\ref{fig:case_study} and \ref{fig:case_study_cond}. In general, \textit{AutoPhrase} can represent meaningful concepts using phrases, but sometimes prone to generate duplicate nodes (\textit{e.g.}, two \textit{``mobile device''} in \textit{AMiner} example). 
\textit{TextRank} select meaningful concepts in word-level which are beneficial for the downstream tasks (\textit{e.g.}, \textit{``beethoven''} in \textit{NYT}, \textit{``mobile''} in \textit{AMiner}, and \textit{``amazing''} in \textit{Yelp}), but the links among the selected concepts are not consistent with the original text. 
The nodes generated from \textit{CMB-MDS} usually contain abundant information but are often in sentence-level, which are not concise and redundant.
\textit{doc2graph} can generate useful concepts with meaningful links, however, the nodes are mainly word-level (\textit{e.g.}, \textit{``mr.''} instead of \textit{``mr. haimovitz''} in \textit{NYT}) and sometimes contain \textit{``\textless unk\textgreater''} or \textit{``-''} which indicate the limitation of this method.
Our approach, \textit{GT-D2G} can represent concepts in both word-level and phrase-level ways which are concise, semantic-rich, and beneficial for downstream tasks (\textit{e.g.}, \textit{``beethoven cello''} in \textit{NYT}).

\subsection{Classification Results (\textit{RQ2})}
\label{ssec:results}

\begin{table}[htbp!]
\caption{Document classification accuracies(\%).}
\centering
\resizebox{\linewidth}{!}{%
\begin{tabular}{cccc}
\toprule
Model & NYT & AMiner & Yelp \\ \midrule
\textit{Bi-LSTM}            & $87.52\pm 3.01$ & $59.32\pm 2.71$  & $78.46 \pm 1.46$    \\
\textit{BERT-base} & $\underline{97.54}\pm0.16$ &$\underline{73.62}\pm0.06$ & $\underline{85.34}\pm0.08$\\ \midrule
\textit{AutoPhrase}         & $92.42\pm	0.65$   & $59.63\pm 0.85$   & $72.66\pm 0.33$     \\
\textit{TextRank}           & $89.48\pm0.07$   & $57.47\pm 0.31$   & $70.25\pm 0.61$     \\
\textit{CMB-MDS}            & $87.68\pm 0.72$   & $51.93\pm 2.02$   & $65.63\pm 2.07$     \\
\textit{doc2graph}    &  $90.81\pm 1.00$  &   $67.06\pm 1.32$  & $79.89 \pm 0.52$     \\ 
\midrule
\textit{GT-D2G-init}   & $93.65\pm0.86$     & $66.76\pm1.77$       & $80.15\pm 0.80$       \\
\textit{GT-D2G-path} &     $95.26\pm 0.13$ &   $68.23\pm 0.23$ &    $80.86\pm 0.97$  \\
%\textbf{GT-D2G-path-var}  & $95.29\pm 0.45$   & $68.25\pm 1.11$   & $80.50\pm 0.83$   \\
\textit{GT-D2G-neigh}             & $95.34\pm 0.33$    & $\mathbf{68.53}\pm 1.02$   & $80.92\pm 0.50$    \\
\textit{GT-D2G-var}        & $\mathbf{95.46}\pm 0.49$    &     $68.37\pm 1.05$ &    $\mathbf{80.98}\pm 0.51$     \\
\bottomrule
\end{tabular}%
}
\label{tab:main_results}
\end{table}

To answer \textit{RQ2}, we conduct the document classification experiments on three text corpora. 
The generated concept maps have $n$ concepts.
To compare our methods with baseline methods conveniently, we set $n=10$ for all graph-based baselines and non-flexible \textit{GT-D2G} variants (\textit{-path} and \textit{-neigh}).
For \textit{GT-D2G-init}, $n$ is equal to the total number of nodes of constructed initial graphs. For the flexible \textit{GT-D2G} variant (\textit{-var}), we set $n\leq 10$.
Table \ref{tab:main_results} shows the classification performance of our methods and the compared methods. 
We observe that \textit{GT-D2G} consistently outperforms all baseline methods except \textit{BERT-base} on all three datasets, which indicates that the integration of semantic-rich initial concept maps from NLP pipelines and graph translation based on the weak supervision in our methods benefit the downstream tasks significantly.
Notably, both \textit{Bi-LSTM} and \textit{BERT-base} are not capable of generating concept maps. As we mentioned before, the goal of \textit{GT-D2G} is not to beat all SOTA document classification methods, but to achieve a competitive performance while providing interpretable structured knowledge representation. Consequently, in the following comparison elaborations, we exclude these two methods when we mention ``baseline methods''.

Compared with traditional graph-based approaches, \textit{GT-D2G} gains 3\%, 15\%, 11\% over the best results of traditional approaches on \textit{NYT}, \textit{AMiner}, and \textit{Yelp}, respectively. Moreover, it surpasses the end-to-end \textit{doc2graph} method by 5\%, 2\% and 1\%, correspondingly. 
As mentioned in the toy example (Fig.~\ref{fig:intro_examples}) and $\S$\ref{ssec:experiment-settings}, both \textit{AutoPhrase}, \textit{TextRank} and \textit{CMB-MDS} are existing unsupervised concept map generation models. These three models are capable of generating concept maps according to their own customized metrics (\eg, frequency-based, connectivity-based, summarization-based), but they can not utilize the downstream task's signals to supervise the generation process. Consequently, concepts generated by these models are not task-oriented, and thus leading to the poor classification performance. On the other hand, \dg is the only compared model that is specifically designed for weakly-supervised concept map generation. As reflected in the experimental results, \dg is the major competitor of our \gtd (excluding the SOTA document classification models).

To better understand the effectiveness of our proposed techniques ($\S$\ref{sec:approach}), we closely study the four variants of \textit{GT-D2G} regarding the effectiveness of NLP pipelines (\textit{-init}), node-and-link iterative generation (\textit{-path} and \textit{-neigh}), and flexible-size graph generation (\textit{-var}).
In particular, to evaluate the effectiveness of incorporating NLP pipelines, we implement \textit{GT-D2G-init} that directly encodes all nodes in the initial semantic-rich concept maps to make predictions. Table~\ref{tab:main_results} show that \textit{GT-D2G-init} outperforms all traditional graph-based baselines with $1.23$ on \textit{NYT}, $7.13\%$ on \textit{AMiner}, and $7.49\%$ on \textit{Yelp}. Comparing \textit{GT-D2G-init} with \dg, \textit{GT-D2G-init} achieves $1.23\%$ and $0.26\%$ gains on \textit{NYT} and \textit{Yelp}, while \textit{GT-D2G-init} is worse by $0.3\%$ on \textit{AMiner}. Hence, the observed experimental results support the benefits of utilizing concept maps derived from NLP pipelines.
Upon \textit{GT-D2G-init}, the other three variants add the Graph Translator module to obtain a more concise concept map, since the initial concept maps often contain 20-40 nodes and the translated concept maps contain less than 10 nodes. 
According to the experimental results, the translated concept maps are preferable to initial concept maps, as they can further improve \textit{GT-D2G-init} by $1.81\%$ on \textit{NYT},  $1.77\%$ on \textit{AMiner}, and $0.83\%$ on \textit{Yelp}.

To explore a proper way to generate edges, we implement and compare two methods, \textit{GT-D2G-path} and \textit{GT-D2G-neigh}. 
\textit{GT-D2G-path} only generates edges based on the relations of concepts in text sequence while \textit{GT-D2G-neigh} links each node with its all possible neighbors.
As shown in Table~\ref{tab:main_results}, \textit{GT-D2G-neigh} is consistently better than \textit{GT-D2G-path} on all three datasets, which well supports our argument that generating edges among all possible neighbors is preferable to generating edges as a sequence of paths starting from the node. Furthermore, \textit{GT-D2G-var} addresses the fixed size issue of \textit{doc\-2graph} and the experiment results of \textit{GT-D2G-var} illustrate the benefits of generating flexible size of concept maps.
More discussion about generating size-flexible concept maps are in $\S$\ref{ssec:flexibility}.

\subsection{Labeling Efficiency Evaluation (\textit{RQ3})}
\label{ssec:few-shot}
\begin{figure*}[tb!]
  \begin{subfigure}[b]{0.31\textwidth}
  \begin{tikzpicture}[scale=0.7]
\begin{axis}[
    xlabel={Training Percentage (\%)},
    ylabel={Accuracy},
    xmin=0, xmax=10,
    ymin=20, ymax=65,
    legend pos=south east,
    legend style={fill=none},
    ymajorgrids=true,
    grid style=dashed,
]
\addplot+[
    thick,
    color=yrbL4,
    mark=x,
    ]
    coordinates {
    (0.1,34.75)(0.25,39.90)(0.5,42.17)(0.75,49.16)(1.0,52.78)(2.5,56.48)(5.0,58.77)(7.5,60.37)(10.0,62.35)
    };
\addplot+[
    thick,
    color=yrbL3,
    mark=o,
    ]
    coordinates {
    (0.1,25.69)(0.25,28.71)(0.5,32.46)(0.75,37.90)(1.0,37.71)(2.5,41.16)(5.0,48.40)(7.5,53.50)(10.0,53.15)
    };
\addplot+[
    thick,
    color=yrbL2,
    mark=diamond,
    ]
    coordinates {
    (0.1,23.10)(0.25,25.68)(0.5,25.76)(0.75,28.17)(1.0,32.49)(2.5,38.85)(5.0,40.62)(7.5,44.77)(10.0,47.40)
    };
\addplot+[
    thick,
    color=yrbL1,
    mark=triangle,
    ]
    coordinates {
    (0.1,28.80)(0.25,32.32)(0.5,35.58)(0.75,38.55)(1.0,41.91)(2.5,47.65)(5.0,48.71)(7.5,51.32)(10.0,52.15)
    };
\addplot+[
    thick,
    color=yrbL0,
    mark=square,
    ]
    coordinates {
    (0.1,28.74)(0.25,31.53)(0.5,36.82)(0.75,39.66)(1.0,41.37)(2.5,45.02)(5.0,47.93)(7.5,49.87)(10.0,50.72)
    };
\legend{GT-D2G,doc2graph,CMB-MDS,TextRank,AutoPhrase}
\end{axis}
\end{tikzpicture}
  \vspace{-0.8cm}
  \caption{NYT}
  \label{tab:nyt_few_shots}
  \end{subfigure}
  \hspace{\fill}
  \begin{subfigure}[b]{0.31\textwidth}
  \begin{tikzpicture}[scale=0.7]
\begin{axis}[
    xlabel={Training Percentage (\%)},
    ylabel={Accuracy},
    xmin=0, xmax=10,
    ymin=20, ymax=65,
    legend pos=south east,
    legend style={fill=none},
    ymajorgrids=true,
    grid style=dashed,
]
\addplot+[
    thick,
    color=yrbL4,
    mark=x,
    ]
    coordinates {
    (0.1,34.75)(0.25,39.90)(0.5,42.17)(0.75,49.16)(1.0,52.78)(2.5,56.48)(5.0,58.77)(7.5,60.37)(10.0,62.35)
    };
\addplot+[
    thick,
    color=yrbL3,
    mark=o,
    ]
    coordinates {
    (0.1,25.69)(0.25,28.71)(0.5,32.46)(0.75,37.90)(1.0,37.71)(2.5,41.16)(5.0,48.40)(7.5,53.50)(10.0,53.15)
    };
\addplot+[
    thick,
    color=yrbL2,
    mark=diamond,
    ]
    coordinates {
    (0.1,23.10)(0.25,25.68)(0.5,25.76)(0.75,28.17)(1.0,32.49)(2.5,38.85)(5.0,40.62)(7.5,44.77)(10.0,47.40)
    };
\addplot+[
    thick,
    color=yrbL1,
    mark=triangle,
    ]
    coordinates {
    (0.1,28.80)(0.25,32.32)(0.5,35.58)(0.75,38.55)(1.0,41.91)(2.5,47.65)(5.0,48.71)(7.5,51.32)(10.0,52.15)
    };
\addplot+[
    thick,
    color=yrbL0,
    mark=square,
    ]
    coordinates {
    (0.1,28.74)(0.25,31.53)(0.5,36.82)(0.75,39.66)(1.0,41.37)(2.5,45.02)(5.0,47.93)(7.5,49.87)(10.0,50.72)
    };
\legend{GT-D2G,doc2graph,CMB-MDS,TextRank,AutoPhrase}
\end{axis}
\end{tikzpicture}
  \vspace{-0.8cm}
  \caption{AMiner}
  \label{tab:aminer_few_shots}
  \end{subfigure}
  \hspace{\fill}
  \begin{subfigure}[b]{0.31\textwidth}
  \begin{tikzpicture}[scale=0.7]
\begin{axis}[
    xlabel={Training Percentage (\%)},
    ylabel={Accuracy},
    xmin=0, xmax=10,
    ymin=60, ymax=80,
    legend pos=south east,
    legend style={fill=none},
    ymajorgrids=true,
    grid style=dashed,
]
\addplot+[
    thick,
    color=yrbL4,
    mark=x,
    ]
    coordinates {
    (0.1,70.86)(0.25,72.24)(0.5,72.70)(0.75,73.22)(1.0,74.48)(2.5,76.50)(5.0,77.59)(7.5,78.01)(10.0,78.10)
    };
\addplot+[
    thick,
    color=yrbL3,
    mark=o,
    ]
    coordinates {
    (0.1,65.12)(0.25,65.09)(0.5,65.58)(0.75,66.13)(1.0,65.65)(2.5,68.02)(5.0,71.01)(7.5,72.57)(10.0,73.29)
    };
\addplot+[
    thick,
    color=yrbL2,
    mark=diamond,
    ]
    coordinates {
    (0.1,63.75)(0.25,63.94)(0.5,63.89)(0.75,63.84)(1.0,63.90)(2.5,63.90)(5.0,63.99)(7.5,64.53)(10.0,64.80)
    };
\addplot+[
    thick,
    color=yrbL1,
    mark=triangle,
    ]
    coordinates {
    (0.1,65.38)(0.25,65.96)(0.5,65.84)(0.75,66.29)(1.0,67.12)(2.5,67.85)(5.0,69.29)(7.5,69.97)(10.0,70.29)
    };
\addplot+[
    thick,
    color=yrbL0,
    mark=square,
    ]
    coordinates {
    (0.1,65.57)(0.25,65.39)(0.5,65.53)(0.75,65.82)(1.0,66.01)(2.5,66.81)(5.0,67.62)(7.5,68.47)(10.0,68.89)
    };
\legend{GT-D2G,doc2graph,CMB-MDS,TextRank,AutoPhrase}
\end{axis}
\end{tikzpicture}
  \vspace{-0.8cm}
  \caption{Yelp}
  \label{tab:yelp_few_shots}
  \end{subfigure}
   \hspace{\fill}
\caption{Test accuracies by varying the proportions of
training data (ranging from 0.1\%, 0.25\%, $\dots$ to 10.00\%).}
\label{fig:few-shots}
\end{figure*}
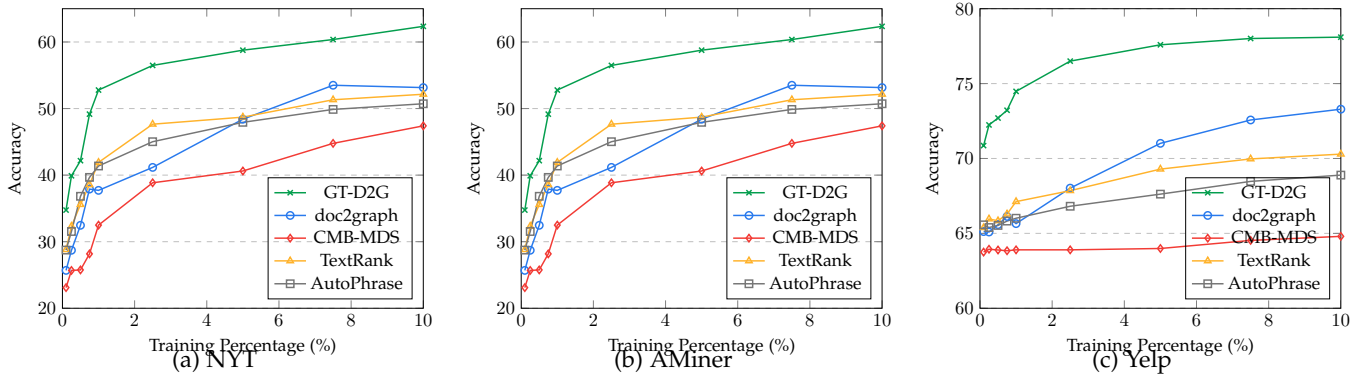

To demonstrate the labeling efficiency of \textit{GT-D2G} over other concept map generation methods, we conduct experiments with different proportions (0.1\%, 0.25\%, 0.50\%, 0.75\%, 1.00\%, 2.50\%, 5.00\%, 7.50\%, 10.00\%) of the training data.  
To get a stable test accuracy, we take the average value among three trials of each experiment by applying different random seeds. 
The average test accuracies of \textit{NYTimes}, \textit{AMiner}, and \textit{Yelp} datasets were shown in Fig.~\ref{fig:few-shots} respectively, which answer \textit{RQ3}. 

We can observe that our approach \textit{GT-D2G} has higher test accuracy than the other approaches from the beginning, with only 0.1\% of the training data. 
In addition, with the increasing of the training data size, our model has steeper growth curves of test accuracy, which shows its effectiveness in exploiting limited supervision, and makes it maintain excellent performance during the whole label efficiency evaluation with limited labeled data. These results demonstrate the labeling efficiency of our model, which is enabled by the semantic-rich initial concept maps ($\S$\ref{ssec:initial-map}) and the Gumbel-softmax training technique ($\S$\ref{ssec:train_tech}).
Therefore, \textit{GT-D2G} can generate concept maps at scales not only without ground-truth training graphs but also without significant amounts of downstream task supervision.

\subsection{Flexibility Evaluation (\textit{RQ4})}
\label{ssec:flexibility}

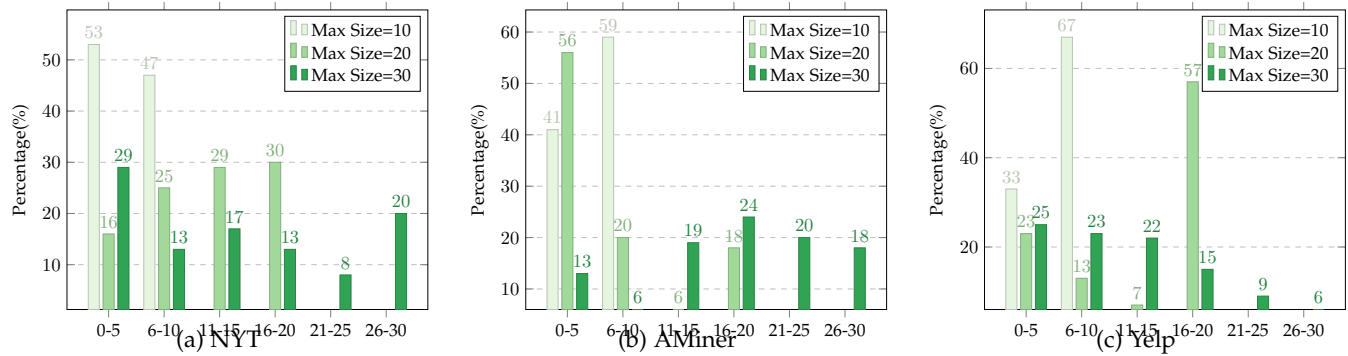
\begin{figure*}[tb!]
\centering
    \begin{subfigure}[!]{0.31\textwidth}
    \begin{tikzpicture}[scale=0.7]
\begin{axis}[
    x tick label style={
		/pgf/number format/1000 sep=},
	symbolic x coords={0-5, 6-10, 11-15, 16-20, 21-25, 26-30},
	ylabel=Percentage(\%),
	enlargelimits=0.15,
	ybar,
	bar width=6pt,
	ymajorgrids=true,
    grid style=dashed,
    nodes near coords,
    nodes near coords align={vertical},
	]
\addplot+ [ylgn1!80!black, fill=ylgn1]
    coordinates{(0-5,53) (6-10,47)};
\addplot+ [ylgn2!80!black, fill=ylgn2]
    coordinates{(0-5,16) (6-10,25) (11-15,29) (16-20,30)};
\addplot+ [ylgn3!80!black, fill=ylgn3]
    coordinates{(0-5,29) (6-10,13) (11-15,17) (16-20,13) (21-25,8) (26-30,20)};
\legend{Max Size=10,Max Size=20,Max Size=30}
\end{axis}
\end{tikzpicture}
%maxlen=10, nlen_entropy=2.62: [(0, 77), (1, 228), (2, 252), (3, 43), (4, 55), (5, 42), (6, 46), (7, 24), (8, 20), (9, 14), (10, 507)]
%maxlen=20,nlen_entropy=4.21: [(0, 39), (1, 20), (2, 28), (3, 31), (4, 48), (5, 42), (6, 58), (7, 57), (8, 59), (9, 75), (10, 83), (11, 99), (12, 79), (13, 67), (14, 75), (15, 59), (16, 68), (17, 61), (18, 43), (19, 31), (20, 186)]
%maxlen=30,nlen_entropy=4.41: [(0, 119), (1, 152), (2, 36), (3, 31), (4, 23), (5, 22), (6, 20), (7, 25), (8, 28), (9, 36), (10, 55), (11, 49), (12, 55), (13, 45),     (14, 39), (15, 32), (16, 34), (17, 39), (18, 37), (19, 37), (20, 26), (21, 29), (22, 28), (23, 18), (24, 20), (25, 11), (26, 9), (27, 9), (28, 8), (29, 7), (30, 229)]
    \vspace{-0.8cm}
    \caption{NYT}
    \end{subfigure}
    \hspace{\fill}
    \begin{subfigure}[!]{0.31\textwidth}
    \begin{tikzpicture}[scale=0.7]
\begin{axis}[
    x tick label style={
		/pgf/number format/1000 sep=},
	symbolic x coords={0-5, 6-10, 11-15, 16-20, 21-25, 26-30},
	ylabel=Percentage(\%),
	enlargelimits=0.15,
	enlarge y limits={upper=0},
	ybar,
	bar width=6pt,
	ymajorgrids=true,
    grid style=dashed,
    nodes near coords,
    nodes near coords align={vertical},
	]
\addplot+ [ylgn1!80!black, fill=ylgn1]
    coordinates{(0-5,41) (6-10,59)};
\addplot+ [ylgn2!80!black, fill=ylgn2]
    coordinates{(0-5,56) (6-10,20) (11-15,6) (16-20,18)};
\addplot+ [ylgn3!80!black, fill=ylgn3]
    coordinates{(0-5,13) (6-10,6) (11-15,19) (16-20,24) (21-25,20) (26-30,18)};
\legend{Max Size=10,Max Size=20,Max Size=30}
\end{axis}
\end{tikzpicture}
%maxlen=10, nlen_entropy=nlen_entropy=2.50: [(0, 109), (1, 35), (2, 248), (3, 78), (4, 215), (5, 196), (6, 130), (7, 46), (8, 28), (9, 20), (10, 1064)]
%maxlen=20, nlen_entropy=3.63: [(0, 30), (1, 353), (2, 201), (3, 250), (4, 219), (5, 157), (6, 112), (7, 96), (8, 92), (9, 70), (10, 62), (11, 41), (12, 30), (13, 23), (14, 15), (15, 13), (16, 10), (17, 11), (18, 10), (19, 2), (20, 372)]
%maxlen=30, nlen_entropy=4.70: [(0, 11), (1, 135), (2, 67), (3, 32), (4, 13), (5, 16), (6, 20), (7, 18), (8, 29), (9, 38), (10, 35), (11, 55), (12, 78), (13, 91), (14, 95), (15, 92), (16, 100), (17, 85), (18, 118), (19, 101), (20, 108), (21, 95), (22, 101), (23, 90), (24, 75), (25, 67), (26, 64), (27, 65), (28, 48), (29, 35), (30, 192)]
    \vspace{-0.8cm}
    \caption{AMiner}
    \end{subfigure}
    \hspace{\fill}
    \begin{subfigure}[!]{0.31\textwidth}
    \begin{tikzpicture}[scale=0.7]
\begin{axis}[
    x tick label style={
		/pgf/number format/1000 sep=},
	symbolic x coords={0-5, 6-10, 11-15, 16-20, 21-25, 26-30},
	ylabel=Percentage(\%),
	enlargelimits=0.15,
	enlarge y limits={upper=0},
	ybar,
	bar width=6pt,
	ymajorgrids=true,
    grid style=dashed,
    nodes near coords,
    nodes near coords align={vertical},
	]
\addplot+ [ylgn1!80!black, fill=ylgn1]
    coordinates{(0-5,33) (6-10,67)};
\addplot+ [ylgn2!80!black, fill=ylgn2]
    coordinates{(0-5,23) (6-10,13) (11-15,7) (16-20,57)};
\addplot+ [ylgn3!80!black, fill=ylgn3]
    coordinates{(0-5,25) (6-10,23) (11-15,22) (16-20,15) (21-25,9) (26-30,6)};
\legend{Max Size=10,Max Size=20,Max Size=30}
\end{axis}
\end{tikzpicture}
%maxlen=10, nlen_entropy=2.82: [(0, 431), (1, 67), (2, 45), (3, 72), (4, 78), (5, 141), (6, 190), (7, 195), (8, 207), (9, 132), (10, 978)]
%maxlen=20, nlen_entropy=2.94: [(0, 66), (1, 127), (2, 144), (3, 68), (4, 79), (5, 99), (6, 82), (7, 78), (8, 62), (9, 49),      (10, 52), (11, 44), (12, 35), (13, 31), (14, 42), (15, 20), (16, 27), (17, 27), (18, 30), (19, 36), (20, 1338)]
%maxlen=30, nlen_entropy=4.77: [(0, 105), (1, 204), (2, 87), (3, 80), (4, 73), (5, 79), (6, 100), (7, 92), (8, 117), (9, 135     ), (10, 127), (11, 135), (12, 129), (13, 110), (14, 104), (15, 90), (16, 98), (17, 87), (18, 74), (19, 61), (20, 50), (21, 61), (22, 56), (23, 46), (24, 40), (25, 31), (26, 34), (27, 20), (28, 19), (29, 22), (30, 70)]
    \vspace{-0.8cm}
    \caption{Yelp}
    \end{subfigure}
    \hspace{\fill}
 \caption{Graph size distributions on different max graph sizes.}    
\label{fig:exp_flexibility}
\end{figure*}

As discussed in $\S$\ref{ssec:results}, the $GT-D2G-var$ variant that is capable of generating flexible sizes of concept maps achieves the best document classification performance on two datasets (NYT and Yelp), while achieving the runner-up on the remaining dataset (AMiner). The observed experimental results justify the importance of the size-flexible property for concept map generation models.

To provide more insights, we further conduct experiments to explore the factors that impact the sizes of generated concept maps. As noted in the Training Techniques ($\S$\ref{ssec:graph-translation}), our framework is able to generate variable sizes of graphs by applying the RBF kernel-based graph size penalty and the content coverage penalty. These two penalties imply a trade-off between conciseness and completeness of generated concept maps.
Fig.~\ref{fig:exp_flexibility} shows the size distribution of the generated graphs on three datasets when the maximum graph size is set to be 10, 20, or 30 nodes. 
As can be seen, our \textit{GT-D2G} can generate graphs with variable sizes as the size distribution varies according to the following two major factors: (a) input text complexity (across three datasets); (b) the preset hyperparameter ``\textit{max size}'' (across different max sizes). For the input text complexity, we know that NYT and AMiner contain rather long and formal news articles and scientific reports, while Yelp contains short and informal online user-generated restaurant reviews. Consequently, concept maps derived from Yelp are inclined to have small sizes, while concept maps from NYT and AMiner have more evenly size distributions (when the \textit{max size} is set to 30). For the hyperparameter \textit{max size}, we can clearly see the set value bounds the actual sizes of generated graphs.

\section{Conclusions}
\label{sec:conclusion}

In this work, we aim to tackle the concept map generation task by graph translation networks.
Without any gold training concept maps, the proposed \gtd framework is able to translate the initial concept maps into the target concise concept maps under the weak supervision from downstream tasks. The quality of generated concept maps is validated through both downstream task performance and human evaluation, in which \gtd outperforms other concept map generation methods by a wide margin.
In the future, we plan to find more meaningful downstream tasks to demonstrate the effectiveness and generalizability of \gtd, and even study it in multi-task settings.

\ifCLASSOPTIONcompsoc
  \section*{Acknowledgments}
\else
  \section*{Acknowledgment}
\fi
This is an extended and revised version of a preliminary conference short paper that was presented in SIGIR 2020~\cite{DBLP:conf/sigir/YangZWL020}. This work significantly expands the weakly supervised concept map generation model, and it provides additional analysis by manual evaluation, more compared methods, efficiency evaluation, and comprehensive ablation studies.
This research is supported by the internal fund and GPU servers provided by the Computer Science Department of Emory University. 
Special thanks to Sophy Huang and Celia Hu for their generous help in the human annotation experiments.

\ifCLASSOPTIONcaptionsoff
  \newpage
\fi



%\bibliographystyle{IEEEtran}
\bibliographystyle{IEEEtran}
\bibliography{ref}

\vspace{-1cm}
\begin{IEEEbiography}[{\includegraphics[width=1in,height=1.25in,clip,keepaspectratio]{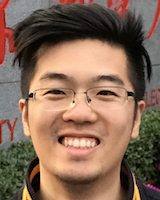}}]{Jiaying Lu}
is a Ph.D. student in Computer Science at Emory University, advised by Prof. Carl Yang. He received his M.S. in Electronic and Communication Engineering and B.S. in Information Engineering at Beijing University of Posts and Telecommunications, China. His research interests include graph data mining, knowledge graph, and multimodal learning.
\end{IEEEbiography}

\vspace{-1cm}
\begin{IEEEbiography}[{\includegraphics[width=1in,height=1.25in,clip,keepaspectratio]{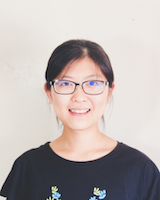}}]{Xiangjue Dong}
is a Ph.D. student in Computer Science and Engineering at Texas A\&M University, advised by Prof. James Caverlee. She earned her M.S. in Computer Science at Emory University in 2021 and M.S. in Civil Engineering at University of Illinois at Urbana-Champaign in 2019. Her research interests lie in natural language processing and conversational AI.
\end{IEEEbiography}

\vspace{-1cm}
\begin{IEEEbiography}[{\includegraphics[width=1in,height=1.25in,clip,keepaspectratio]{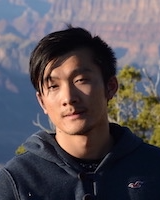}}]{Carl Yang}
is an Assistant Professor of Computer Science in Emory University. He received his Ph.D. in Computer Science at University of Illinois, Urbana-Champaign in 2020, and B.Eng. in Computer Science and Engineering at Zhejiang University in 2014. His research interests span graph data mining, applied machine learning and structured information systems, with applications in knowledge graphs, recommender systems, biomedical informatics and healthcare. Carl's research results have led to over 50 publications in top peer-reviewed journals and conference proceedings including TKDE, KDD, WWW, NeurIPS, ICML, ICDE and SIGIR. He also received the Dissertation Completion Fellowship of UIUC in 2020 and the Best Paper Award of ICDM in 2020.
\end{IEEEbiography}

\end{document}